\documentclass[10pt,twocolumn,letterpaper]{article}

\usepackage[pagenumbers]{cvpr} 

\usepackage{amsmath, amssymb} 
\usepackage{bbm}
\usepackage{multirow}

\usepackage[most]{tcolorbox}

\newtcbox{\actionchip}{on line, box align=base,
  colback=blue!10, colframe=white,
  boxsep=1pt, left=2pt, right=2pt, top=0.5pt, bottom=0.5pt,
  arc=3pt}

\newtcbox{\thingchip}{on line, box align=base,
  colback=red!10, colframe=white,
  boxsep=1pt, left=2pt, right=2pt, top=0.5pt, bottom=0.5pt,
  arc=3pt}

\newcommand{\jy}[1]{{\color{magenta} #1}}

\usepackage{amsmath, amssymb} 
\usepackage{bbm}
\usepackage{multirow}
\usepackage{tabularx,ragged2e}
\newcolumntype{C}{>{\arraybackslash}X}
\usepackage{float}
\usepackage{longtable}

\definecolor{cvprblue}{rgb}{0.21,0.49,0.74}
\usepackage[pagebackref,breaklinks,colorlinks,allcolors=cvprblue]{hyperref}

\def\paperID{38943} 
\def\confName{CVPR}
\def\confYear{2026}

\title{CREward: A Type-Specific Creativity Reward Model}

\author{
Jiyeon Han $^{1}$,\;  Ali Mahdavi-Amiri $^{1}$,\; Hao Zhang $^{1}$,\; and Haedong Jeong $^{2}$\\
$^1$Simon Fraser University \quad $^2$ Sogang University \\
{\tt\small {\{jiyeonh,ali\_mahdavi-amiri,haoz\}@sfu.ca, haedong@sogang.ac.kr}
}
}
\begin{document}
\maketitle
\begin{abstract}
Creativity is a complex phenomenon. When it comes to representing and assessing creativity, treating it as a single undifferentiated quantity would appear naive and underwhelming.
In this work, we learn the \emph{first type-specific creativity reward model}, coined CREward, which spans three creativity ``axes,'' \textit{geometry}, \textit{material}, and \textit{texture}, to allow us to view creativity through the lens of the image formation pipeline.
To build our reward model, we first conduct a human benchmark evaluation to capture human perception of creativity for each type across various creative images. We then analyze the correlation between human judgments and predictions by large vision-language models (LVLMs), confirming that LVLMs exhibit strong alignment with human perception. Building on this observation, we collect LVLM-generated labels to train our CREward model that is applicable to both evaluation and generation of creative images. We explore three applications of CREward: creativity assessment, explainable creativity, and creative sample acquisition for both human design inspiration and guiding creative generation through low-rank adaptation. Code and dataset are available at \url{https://han-j-y.github.io/creward_prj/}.

\end{abstract}    
\section{Introduction}
\begin{figure}
    \centering
    \includegraphics[width=\linewidth]{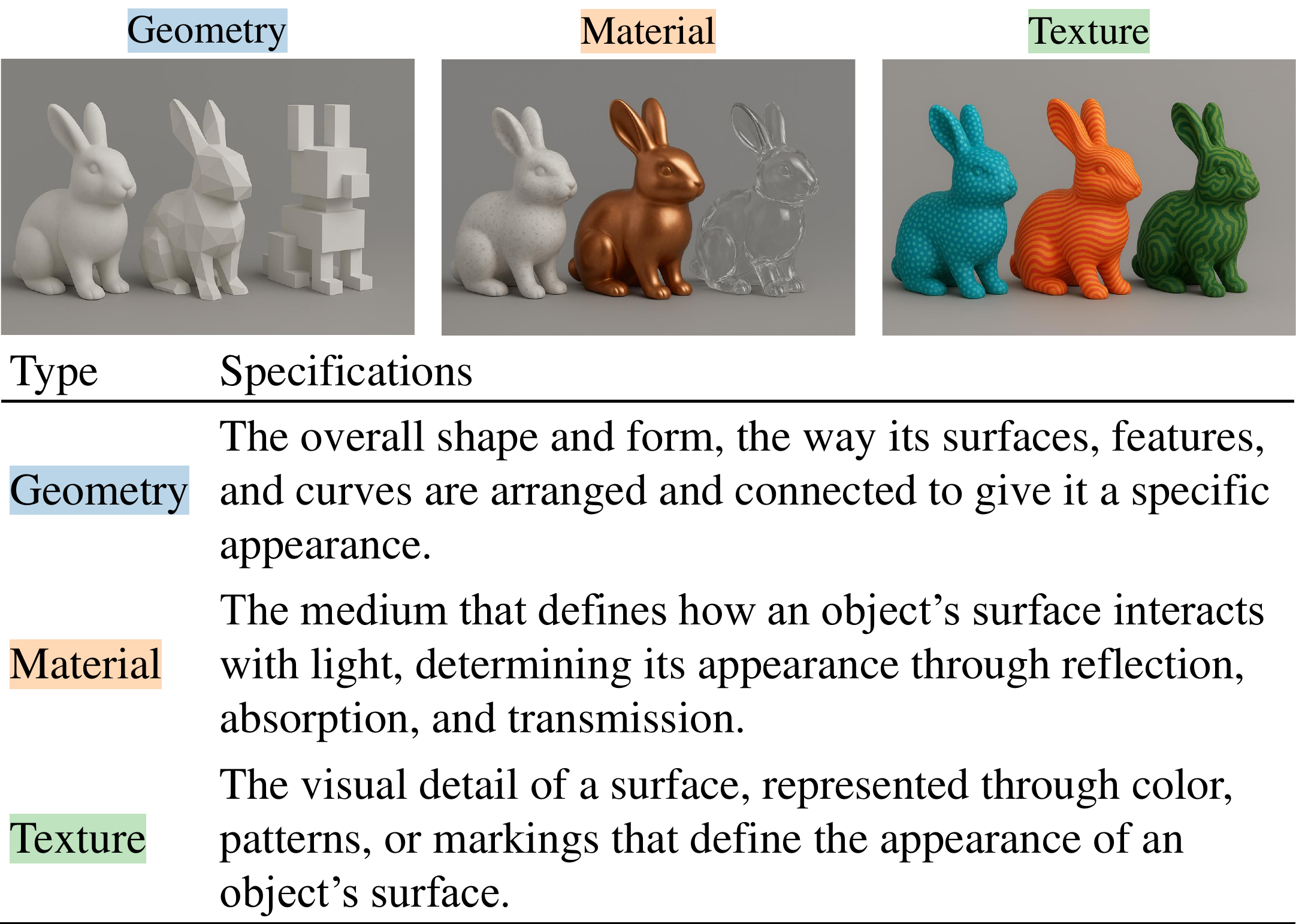}
    \caption{
    Three types of creativity grounded in 3D rendering.
    }
    \label{fig:types}
\end{figure}

\begin{figure*}[t!]
    \centering
    \includegraphics[width=1\linewidth]{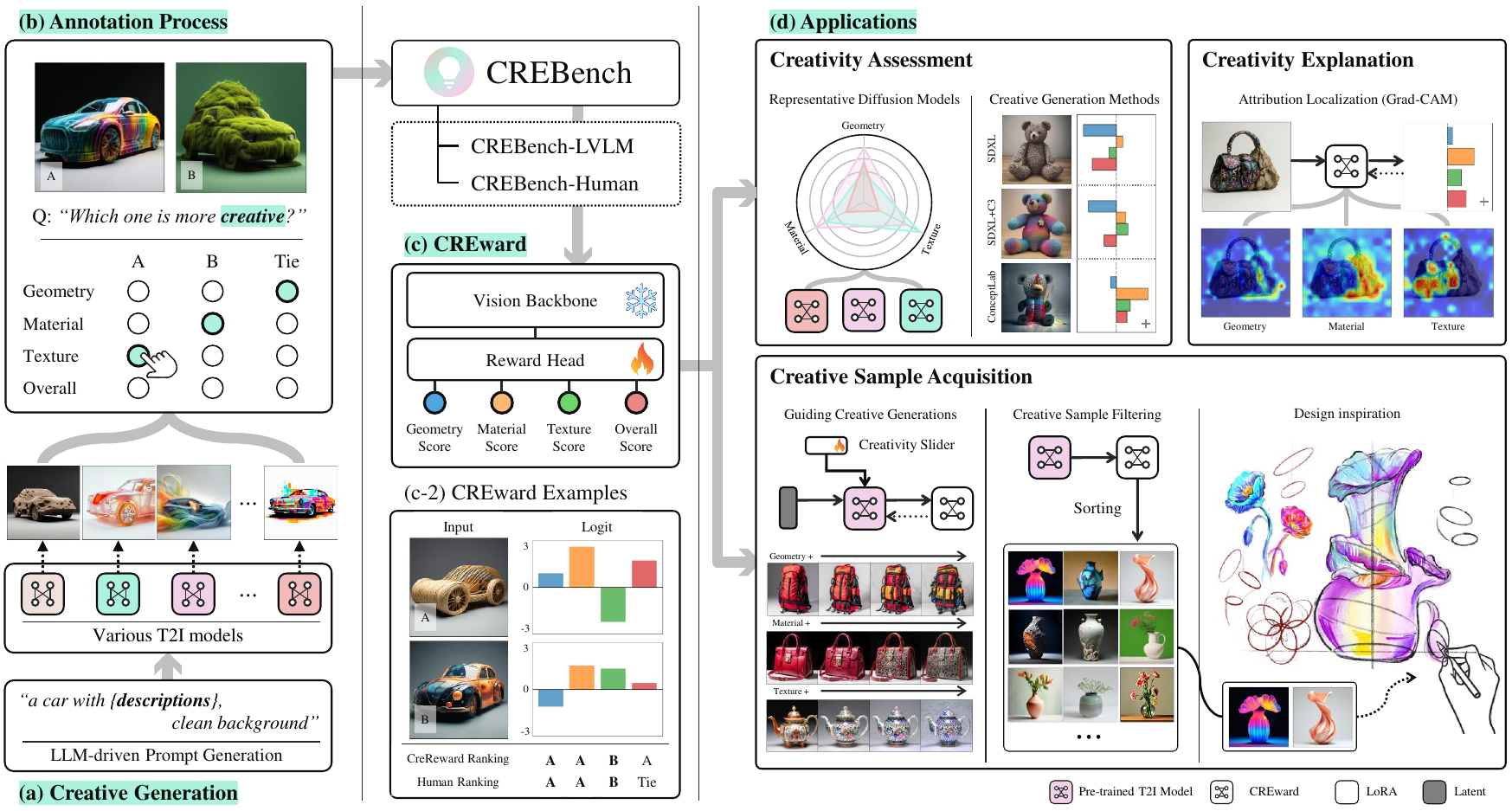}
    \caption{Overview of CreBench, CREward, and their applications.
    (a) LLM-driven prompts and various T2I models are used to collect creative generations.
    (b) Pairwise rankings are collected on randomly sampled image pairs along four types—geometry, material, texture, and overall—using either human annotators or a LVLM; together these comprise CreBench.
    (c) To distill LVLM judgments, we train a lightweight regressor (frozen vision backbone + MLP head) to predict type-wise scores, yielding CREward.
    (d) Applications enabled by CREward: Creativity Assessment, creativity explanation, and creativity sample acquisition.} 
    \label{fig:method_overview}
\end{figure*}

\label{sec:intro}

Generating diverse and creative imagery of common objects can be a fascinating and rewarding endeavor, not only for professional artists but also for casual, and even novice, users. Whether for evaluation, guided exploration, or drawing samples from the creativity spaces for generative modeling, a critical requirement, and also a key challenge, is to define creativity in a precise, quantifiable, and \emph{explainable} way. To address this challenge, we observe that creativity in visual design often manifests itself along interpretable dimensions such as geometry, material, and texture as specified in Figure~\ref{fig:types}. A study of these three creativity types also allows us to view creativity through the lens of the \emph{image formation} pipeline.


Modeling and reasoning about \emph{type-specific} creativity offer several advantages including finer-grained and interpretable evaluation, controllability over creative directions, improved diversity by encouraging exploration across multiple creative dimensions, and stronger alignment with human perception, which is known to differentiate and hierarchically process distinct visual attributes. These considerations motivate the need for frameworks that explicitly represent and assess creativity across multiple types rather than treating creativity as a single undifferentiated quantity.

In this paper, we wish to learn the \emph{first} type-specific \emph{creativity reward model} that is applicable to both evaluation and generation of creative images. As shown in Figure~\ref{fig:method_overview}, we accomplish this goal through several steps. First, we establish a human benchmark, where expert annotators compare pairs of images to judge their creativity with respect to geometry, material, and texture. 
We then assess how well large vision–language models (LVLMs) align with human perception by prompting Gemini-2.5-Flash (Gemini-2.5) and Gemma-3-27B-it (Gemma-3) using the same instructions as the human study. Both models exhibit notable correlation with human judgments, with Gemini-2.5 achieving the highest alignment. These results motivate us to leverage LVLM-generated labels as a scalable alternative to creativity assessment by humans.

Building on this insight, we synthesize a large dataset of creative images using multiple state-of-the-art diffusion models, guided by LLM-generated prompts targeted at each creativity type. We then collect LVLM preference labels for thousands of image pairs and train \emph{CREward}, a reward model that predicts creativity scores. After identifying the best-performing backbone, we evaluate CREward on the human benchmark. Despite being trained solely on LVLM-generated labels, CREward achieves strong alignment with human perception, 
suggesting that LVLM-derived supervision can effectively model creativity and support controllable creative generation.

CREward enables several applications in creativity assessment, generation, and analysis. As a reward model, it provides type-specific creativity scores that allow analyzing generative model's creative capability. CREward also supports creative sample acquisition by filtering large pools of generations to surface the most creative examples for each type, which can then be used for design inspiration in human–AI co-creation workflows. This capability reduces the search burden for designers and helps spark new ideas, as demonstrated in our collaboration with an expert specialized in industrial design; see Figure~\ref{fig:method_overview} (d). CREward further enables guided creative generation, where its scores drive LoRA-based sliders that selectively amplify geometry, material, or texture creativity during denoising. Despite minor entanglements across sliders that we could observe, they effectively guide generations toward the desired creativity type. Finally, CREward integrates naturally with XAI techniques such as Grad-CAM, offering interpretable visualizations of image regions contributing to each creativity type and paving the way for explainable and controllable creative AI systems.

\section{Related Work}

\noindent\textbf{Creativity Assessments.}
While novelty and value are recognized as central to computational creativity \cite{boden1998creativity},
quantitative assessment remains insufficiently explored. Prior works on creative generation~\cite{richardson2024conceptlab,han2025enhancing,feng2025redefining} often rely on costly human evaluation or repurpose generic vision metrics (e.g., FID~\cite{heusel2017gans}, CLIPScore~\cite{hessel2021clipscore}, and Improved Precision and Recall~\cite{kynkaanniemi2019improved}), which are not designed to capture creative qualities.
Rarity Score ~\cite{han2022rarity} measures uncommonness relative to a training distribution, but is constrained by data bias.
\citet{wang2025evaluation} operationalize Boden’s criteria (novelty, value, surprise)~\cite{boden1998creativity}, yet only novelty can be computed at the individual-sample level.
Overall, existing approaches treat creativity as a monolithic notion and offer limited guidance for generative models.

\noindent\textbf{Creative Generation.}
Creative generation has gained substantial attention as the capabilities of generative models have advanced dramatically. 
ConceptLab~\cite{richardson2024conceptlab} achieves new creative sub-concepts by contrastively exploring the concept space to find a new sub-concept that does not exist in the sub-concepts yet while it is still regarded as the target concept. 
\citet{han2025enhancing} observe limited creative generation from specific set of text-to-image generative diffusion models and enhances creative generations by amplifying and manipulating inner features of pretrained models. 
Another line of research focuses on compositional creativity~\cite{feng2025redefining,sun2025creative}, where surprise emerges from combining familiar concepts in unexpected ways. Furthermore, a recent study proposes an agent-based self-refinement framework for creative generation~\cite{venkatesh2025crea}. Nevertheless, controllability over different creative visual aspects is still largely underexplored.

\noindent\textbf{LVLM-based Evaluation.}
Recent large vision–language models (LVLMs) \cite{team2025gemma,comanici2025gemini,qwen3technicalreport} have achieved strong performance across image understanding tasks. In particular, LVLMs provide human-like assessments in visual question answering (VQA) \cite{li2023blip,chen2024mllm} and video question answering \cite{li2023vlm}, demonstrating broad contextual and semantic understanding of visual inputs.
However, it remains unclear whether LVLMs can reliably recognize or evaluate creativity.
In this work, we examine the suitability of LVLMs for creativity assessment by analyzing their alignment with human judgments across fine-grained creativity types.



\noindent\textbf{Reward Models.}
Reward models learn to approximate human preferences and provide effective feedback for generative models. ImageReward~\cite{xu2023imagereward} trains a text–image reward model using large-scale human annotations of alignment and fidelity. VisionReward~\cite{xu2024visionreward} extends this to images and videos through VQA-based preference prediction. UnifiedReward~\cite{wang2025unified} unifies pointwise scoring and pairwise ranking, optionally generating justifications.
Incorporating these reward models into the training loop provides human-aligned supervisory signals, improving a model’s ability to learn preferences consistent with human judgment.

\section{Methodology}
\label{sec:methodology}
We focus on \textit{Geometry}, \textit{Material}, and \textit{Texture} as the primary factors for assessing and analyzing creativity grounded in the 3D rendering pipeline. Their specifications are illustrated in Figure~\ref{fig:types}, with additional discussion of type identification provided in Appendix~\ref{sec:sup_types}.
We first conduct a human benchmark evaluation to capture human perception of creativity for each type across various creative images. We then analyze the correlation between human judgments and LVLM predictions, confirming that LVLMs exhibit strong alignment with human perception.
Building on this observation, we collect LVLM-generated labels to train a reward model to assess and guide creative generation. Figure~\ref{fig:method_overview} illustrates the overview of our methodology in detail.

\begin{table*}[t!]
  \caption{Rank correlation with human ranks. Numbers in parentheses indicate standard deviations over the objects. Boldface indicates the highest score, and underline indicates the second highest score.
  The \actionchip{blue model} denotes the open-source LVLM and \thingchip{red model} denotes the closed source LVLM.}
  \label{tab:quantitative}
  \centering
  \begin{tabular}{@{}c|c|c|c|c|c|c@{}}
    \toprule
    Type & Metric ($\uparrow$) & Inter-Human & CREward \textbf{(ours)} & \thingchip{Gemini-2.5 Flash} & \actionchip{Gemma-3-27b-it} & Surprise~\cite{wang2025evaluation}\\
    \midrule
    \multirow{2}{*}{Geometry} & Rank Corr. & 0.71 (0.07) & \underline{0.59 (0.09)} & \textbf{0.80 (0.02)} & 0.56 (0.11) & 0.42 (0.16) \\
    & Acc. (\%) & -  & \underline{0.72 (0.03)} & \textbf{0.87 (0.06)} & 0.71 (0.06) & 0.66 (0.06) \\
    \midrule
    
    \multirow{2}{*}{Material} & Rank Corr. & 0.63 (0.10) & \underline{0.72 (0.06)} & \textbf{0.75 (0.09)} & 0.66 (0.10) & 0.51 (0.14) \\
    & Acc. (\%)  & -  & \underline{0.77 (0.04)} & \textbf{0.84 (0.07)} & 0.65 (0.05) & 0.71 (0.06) \\
    \midrule
    
    \multirow{2}{*}{Texture} & Rank Corr. & 0.46 (0.10) & \textbf{0.76 (0.12)} & \underline{0.74 (0.06)} & 0.60 (0.05) & 0.49 (0.19) \\
    & Acc. (\%) & -  & \textbf{0.74 (0.04)} & \underline{0.73 (0.04)} & 0.64 (0.03) & 0.66 (0.04) \\
    \midrule
    
    \multirow{2}{*}{Overall} & Rank Corr. & 0.65 (0.10) & \underline{0.61 (0.06)} & \textbf{0.68 (0.04)} & 0.57 (0.08) & 0.56 (0.15) \\
    & Acc. (\%) & -  & \underline{0.72 (0.02)} & \textbf{0.74 (0.03)} & \underline{0.72 (0.03)} & 0.69 (0.04) \\
    
    \bottomrule
  \end{tabular}
\end{table*}

\begin{figure}
    \centering
    \includegraphics[width=1\linewidth]{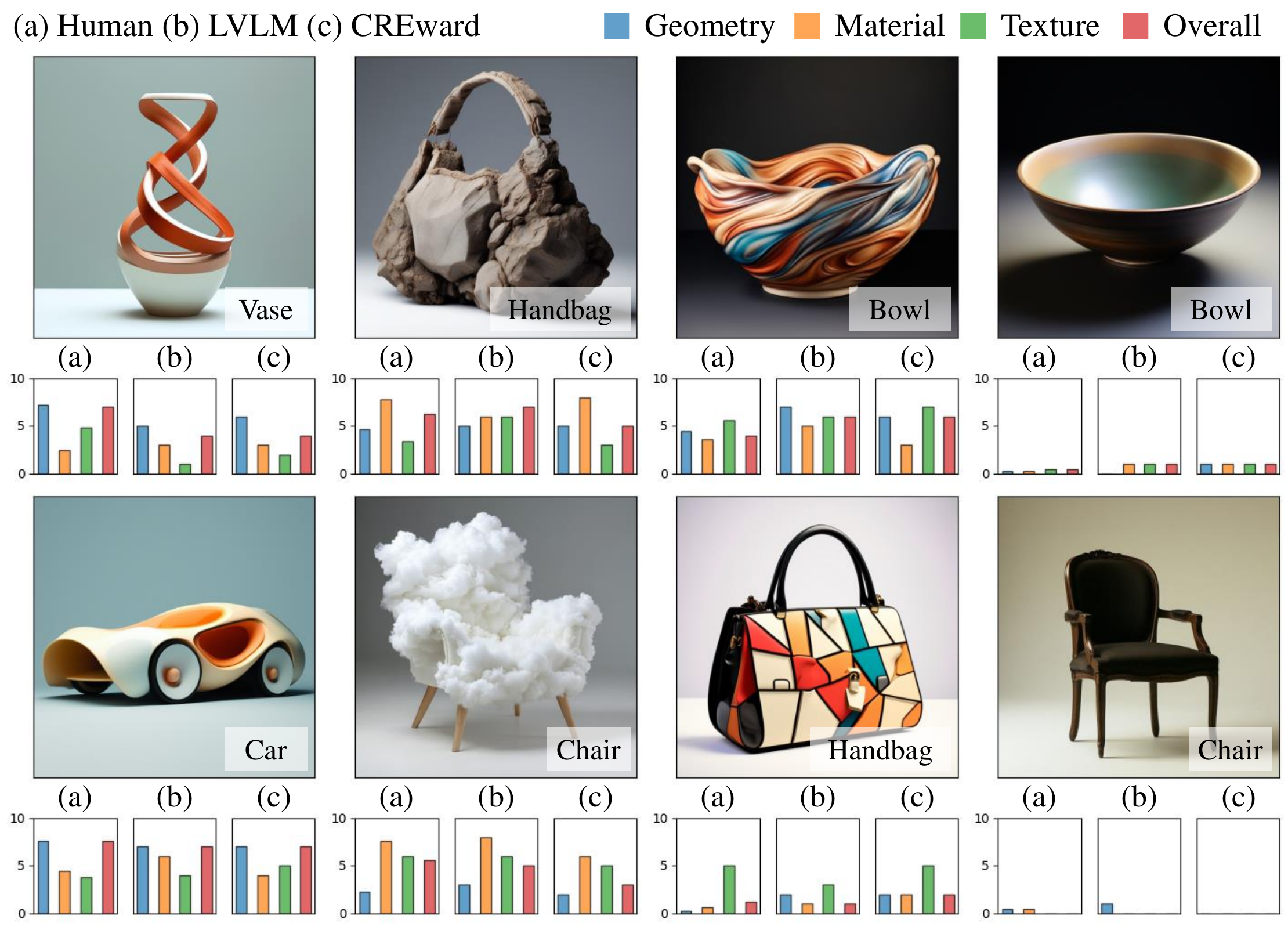}
    \caption{Winning rates ($\uparrow$) derived from preference labels on the benchmark dataset for human evaluation, LLM (Gemini-2.5), and our CREward.}
    \label{fig:assessment_ex}
\end{figure}

\subsection{Human Benchmark}
\label{sec:human_benchmark}
We first conduct a human evaluation of creativity on a small benchmark dataset. Five annotators with design expertise participated in the study. All of them have more than four years of design training, ranging from PhD students to practitioners with up to nine years of industry experience. Since assigning absolute scores can be inconsistent even for a single annotator, we instead ask annotators to perform pairwise comparisons between image pairs. To construct the benchmark dataset, we first generate 5,000 creative images per object using generative models and manually select 20 images that represent diverse levels of creativity across different types. Additionally, we include 5 normal images generated with the prompt ``\textit{a/an \{obj\}}'' as the least creative samples, resulting in a total of 25 images per object.
We conduct the human evaluation on five objects: \textit{chair, car, handbag, bowl,} and \textit{vase}.

We randomly sample 100 image pairs, ensuring that each image appears exactly 8 times. Before the ranking process begins, we provide annotators with detailed descriptions (as shown in Figure~\ref{fig:types}) to aid their understanding of each creativity type. For each pair, annotators are asked to choose the image that appears more creative for a given type or to select \textit{Tie} if they cannot decide, as illustrated in Figure~\ref{fig:method_overview} (b). We additionally ask annotators to evaluate overall creativity to analyze its correlation with each creativity type.
Given the preference labels from the annotators, the winning rate for each of the 25 images is computed as the proportion of pairwise comparisons in which the image is preferred, with normalization by the total number of appearances (8 in our case). We systematically break ties, by assigning `win' to the sample with higher winning rate. 
Then we rank the samples according to the winning rate.
The Inter-Human column in Table~\ref{tab:quantitative} shows the mean Spearman’s rank correlation among annotators, averaged over the five objects. We observe that geometry-based creativity yields higher agreement, whereas texture-based creativity appears to be more subjective.

\subsection{Human--LVLM Creativity Correlation}
\label{sec:human-lvlm correlation}
As current LVLMs offer great performance across various domains, including evaluation and feedback generation, we investigate their reliability in assessing creativity. We employ Gemini-2.5, a closed source LVLM, and Gemma-3, an open-weight model, as our LVLM annotators.
We provide both models with the same descriptions and instructions used in the human evaluation, and post-process their responses to derive rankings same way as for human annotators. To aggregate human annotations, we average the winning rates across annotators and then compute the Spearman rank correlation with human average and LVLM ranking.
We also calculate the preference agreement over 100 pairs, excluding ties from human annotators, and report the results in the Acc. rows of Table~\ref{tab:quantitative}.

The results are notable: Gemini-2.5 shows a high correlation with human perception of creativity, even exceeding the inter-human correlation. However, as a closed-source model, it is less suitable for large-scale assessments due to cost and accessibility constraints. Gemma-3, while not as strong as Gemini-2.5, performs reasonably well and achieves comparable or even higher correlation than human annotators for certain creativity types, such as material and texture.
Inspired by these findings, we propose a creative reward model, \textit{\textbf{CREward}}, which leverages LVLM-generated creativity preference data to emulate human-perception of creativity for each type.

\begin{table}[t!]
    \centering
    \caption{Test preference accuracy for various vision backbones.}
    \label{tab:llm_performance_simplified}\label{tab:vision_backbones}
    \begin{tabular}{lccccc}
        \toprule
        \textbf{Model} & \textbf{\# Params} & \textbf{Geo.} & \textbf{Mat.} & \textbf{Tex.} & \textbf{Ove.} \\
        \hline
        VGG16 & 138M & 0.72 & 0.72 & 0.75 & 0.74 \\
        CLIP & 304M & 0.80 & 0.77 & \textbf{0.80} & 0.78 \\
        DreamSim & 267M & 0.76 & 0.73 & 0.79 & 0.78 \\
        DINOv3 & 824M & 0.78 & 0.70 & 0.76 & 0.78 \\
        SigLIP & 422M & \textbf{0.81} & \textbf{0.78} & \textbf{0.80} & \textbf{0.82} \\
        \bottomrule
    \end{tabular}
\end{table}

\subsection{Collecting LVLM-generated Creativity Ranks}
Since it is challenging to collect real-world images that exhibit specific types of creativity, we synthesize images corresponding to each targeted creativity type. To minimize potential bias from any single generative model, we employ a diverse set of recent text-to-image (T2I) models with varying architectures and training datasets, including Hunyuan-DiT \cite{li2024hunyuandit}, PixArt \cite{chen2023pixartalpha}, Kandinsky v3 \cite{vladimir-etal-2024-kandinsky}, Stable Diffusion v3.5 Large \cite{esser2024scaling}, and Flux-schnell \cite{flux2024}.

To obtain diverse creative outputs, we employ an LLM (ChatGPT-5) to generate 20 prompts per creativity type.
 Among these, 8 are object-agnostic templates, while the remaining 12 are object-specific. Example prompts would be ``\emph{a/an \{obj\} with a creative jelly-like material}'' and ``\emph{a chair with asymmetrical legs and a twisted seat structure}''. We explicitly add ``\emph{clean background}'' to each prompt to ensure the focus remains on the object rather than its background.
 For each prompt, we generate 10 images per model.


Next, we construct training pairs for the reward model using the creative generations. We randomly select two prompts out of the 60, and for each selected prompt, randomly sample one generated image. In total, we collect 1,000 pairs per object, resulting in 5,000 pairs for comparison across all objects.

To obtain LVLM-generated preference labels, we input each image pair into the Gemma-3 model, an open-weight LVLM that shows a high correlation with human judgments of creativity. We provide the model with detailed instructions, including the definitions of each creativity type and the expected answer format, and obtain preference labels for each creativity type as well as an overall preference. An example of LVLM queries is provided in Appendix~\ref{sec:sup_lvlm_anno}.

\subsection{CREward}
To reduce the high computational cost of LVLM-driven creativity evaluation and to improve adaptability to downstream tasks (e.g., reward-based filtering), we adopt a reward model (RM) training strategy inspired by prior work \cite{stiennon2020learning,ouyang2022training,xu2023imagereward}.
Given two images, $x_A$ and $x_B$, ranked across the three creativity types as well as the overall creativity, we formulate preference learning as a binary classification task on triplets, $(x_A,x_B,y)$, where $y\in\{+1,-1,0\}$ denotes whether $x_A$ is preferred ($+1$), $x_B$ is preferred ($-1$), or it is a tie ($0$). We exclude ties during training using a binary mask $m(y)=\mathbbm{1}[y\neq0]$. For type $c$, let $f_\theta^{(c)}$ be the scalar reward model. The pairwise logistic score is defined as,
\begin{align}
    &v^{(c)}(x_A,x_B,y)=\sigma(y\cdot[f^c_\theta(x_A)-f^c_\theta(x_B)]),\\
    &L(\theta, c)=-\mathbb{E}_{(x_A,x_B,y) \sim D}[\log(m(y)v^{(c)}(x_A,x_B,y)],
\end{align}
where $\sigma(\cdot)$ is sigmoid.
We evaluate multiple vision backbones\footnote{VGG16 (pre-trained on ImageNet) \cite{simonyan2014very}; CLIP (ViT-L/14) \cite{radford2021learning}; DreamSim (ensemble mode) \cite{fu2023dreamsim}; DINOv3 (vith16plus-pretrain-lvd1689m) \cite{simeoni2025dinov3}; SigLIP (Gemma-3 vision encoder) \cite{team2025gemma}.} and attach a 5-layer MLP head (with ReLU activations, dropout $p=0.2$) that outputs four scalar scores—one per each type in addition to \textit{overall}. We use SDXL-DMD2~\cite{yin2024improved,yin2024onestep} as our base model with TAESD\footnote{\url{https://github.com/madebyollin/taesd}}, which benefits from a lightweight distilled vision decoder.
During training, the backbone is frozen and only the MLP head is optimized for 20 epochs using a fixed train–val–test split. For each backbone, we select the best checkpoint by validation accuracy, then report test preference accuracy and choose the final backbone for our RM accordingly. Table~\ref{tab:vision_backbones} summarizes test preference accuracy across backbones. Based on these results, we adopt the SigLIP vision encoder as our backbone and refer to the resulting reward model as CREward.

\begin{table}
  \caption{Rank correlation between type-specific creativity and overall creativity, showing the relative influence of each type across different annotation methods.
Numbers in parentheses indicate standard deviation over the objects.}
  \label{tab:type_corr}
  \centering
  \begin{tabular}{@{}r@{\hspace{3mm}}c@{\hspace{3mm}}c@{\hspace{3mm}}c@{\hspace{3mm}}c@{}}
    \toprule
    \textbf{Type} & \textbf{Human} & \textbf{Gemini-2.5} & \textbf{Gemma-3} & \textbf{CREward}\\
    \midrule
    Geo. & 0.84 (0.07) & 0.96 (0.03) & 0.75 (0.10) & 0.80 (0.06) \\
    Mat. & 0.65 (0.08) & 0.82 (0.08) & 0.67 (0.19) & 0.44 (0.09) \\
    Tex. & 0.58 (0.13) & 0.71 (0.07) & 0.69 (0.11) & 0.39 (0.15) \\
    \bottomrule
  \end{tabular}
\end{table}

\subsection{Alignment with Human Creativity Perception}
To validate the trained CREward, we compute its rank correlation and preference accuracy against human judgments on the CREBench-Human dataset. For each image pair, we obtain type-specific scores from CREward and assign the preference to the image with the higher score. Using these preferences, we calculate the rank correlation and preference accuracy following the procedure described in Section~\ref{sec:human-lvlm correlation}. We use the \textit{Surprise} score \cite{wang2025evaluation} as a baseline computational creativity metric.
Given that Surprise does not distinguish between creativity types, we compute rank correlations between our type-specific ranks and its type-agnostic rankings. The results are summarized in Table~\ref{tab:quantitative}.

\begin{figure}[t!]
    \centering
    \includegraphics[width=1\linewidth]{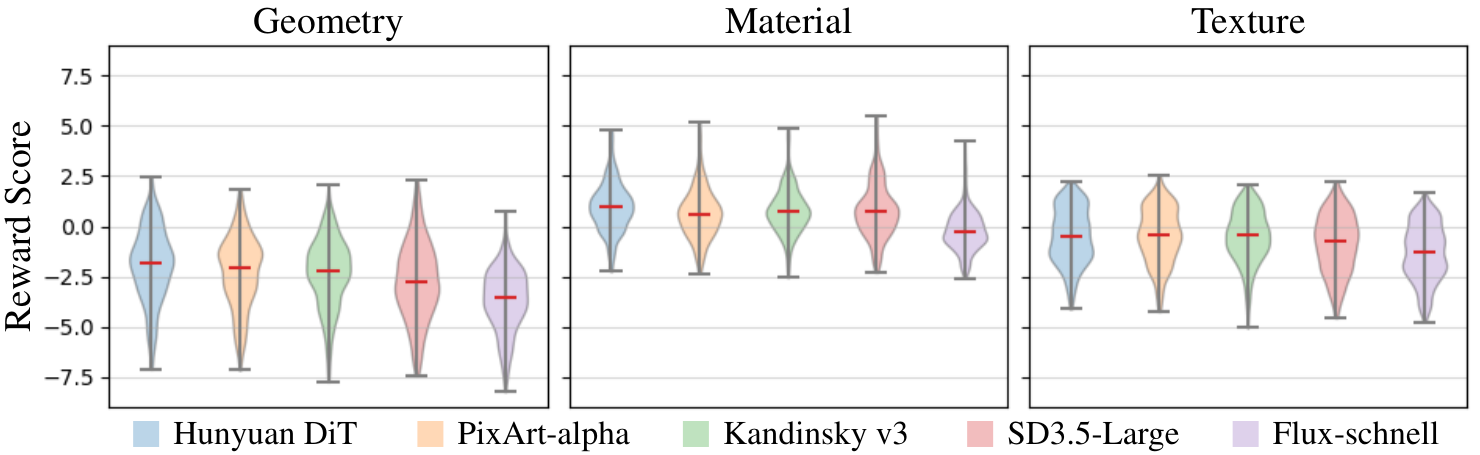}
    \caption{Violin plots of reward distributions for creative generations from various representative diffusion models.}
    \label{fig:model_comparison}
\end{figure}

Although CREward is trained solely on labels generated from Gemma-3, it achieves the second-highest rank correlation with human perception, surpassing even the closed-source Gemini-2.5 on texture creativity.
Figure~\ref{fig:assessment_ex} provides examples from CreBench-Human along with winning rates derived from human annotators, Gemini-2.5, and CREward. These examples demonstrate that CREward effectively captures relative creativity, producing rankings that align well with human judgments and effectively distinguish different creativity types.

\begin{figure*}
    \centering
    \includegraphics[width=\linewidth]{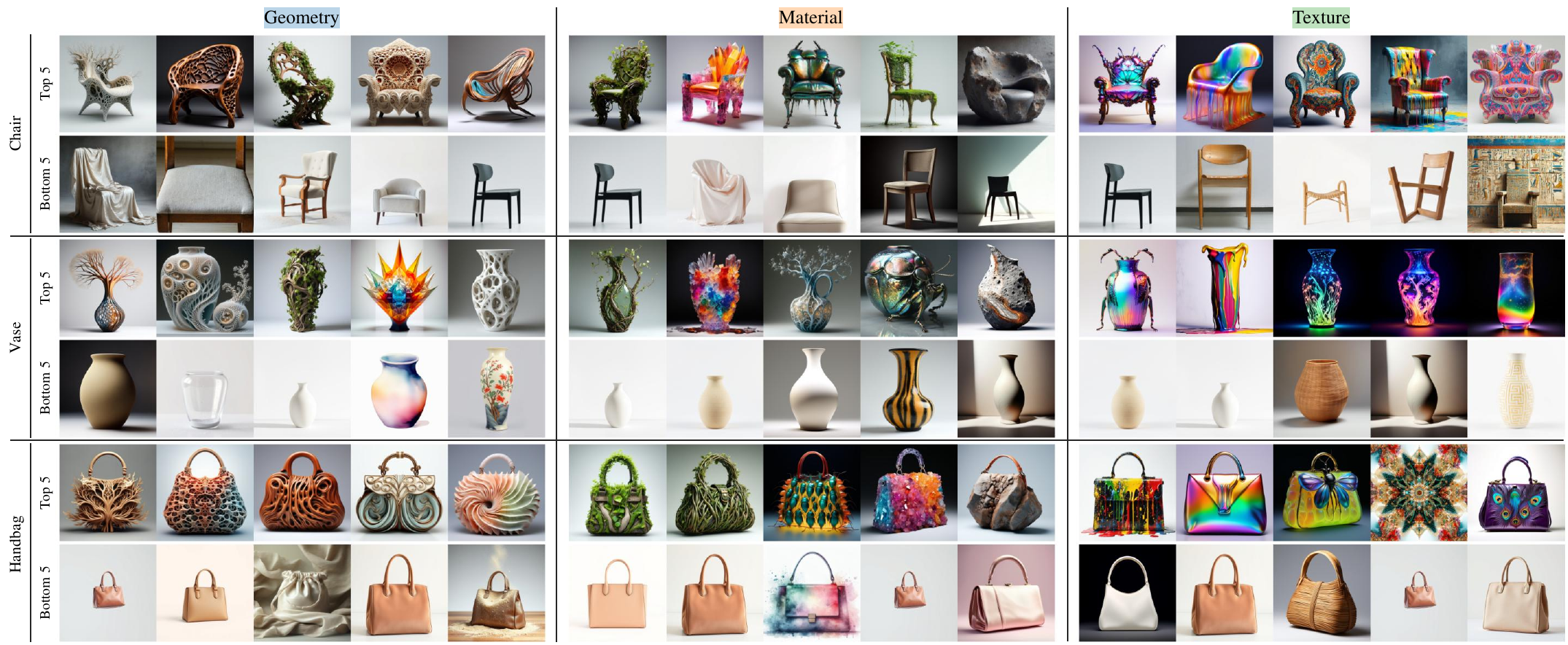}
    \caption{Top/Bottom 5 ranked generations for each creativity type from 100 LLM-generated (type-agnostic) creative prompts .}
    \label{fig:cre_filtering}
\end{figure*}

\begin{figure}[t!]
    \centering
    \includegraphics[width=1\linewidth]{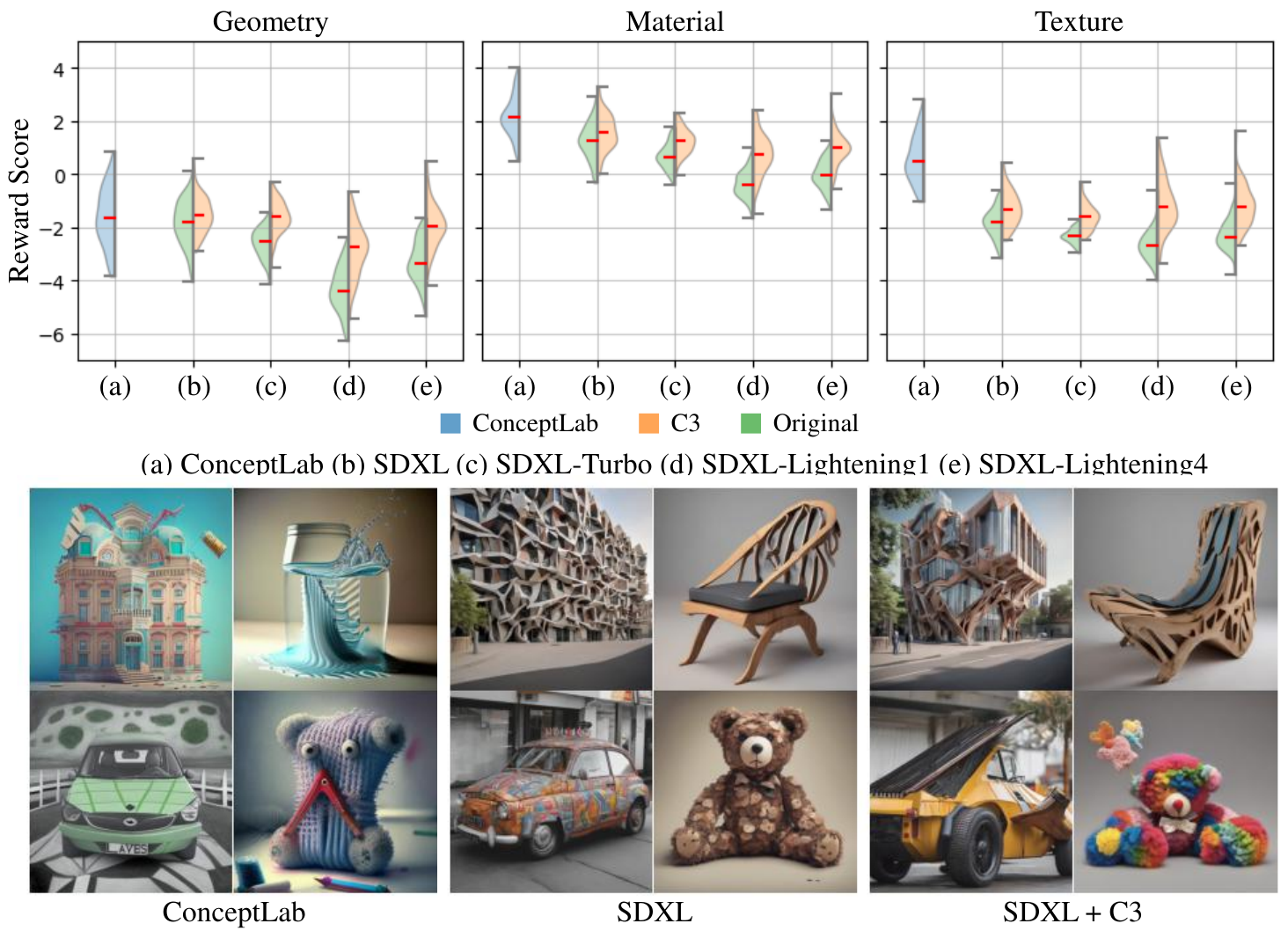}
    \caption{Violin plots of reward distributions for ConceptLab and C3, where C3 is compared with its original counterpart. Default option is used in ConceptLab and ``\textit{a creative {obj}}" is used for C3 and original counterpart.}
    \label{fig:cre_method}
\end{figure}

\subsection{Type-wise Correlation with Overall Creativity}

Based on the collected preferences and rankings in the benchmark dataset, we analyze which types of creativity most strongly influence the overall perception of creativity. Table~\ref{tab:type_corr} reports the Spearman rank correlations between the type-specific rankings and the overall creativity ranking.

Across human annotators, LVLMs, and CREward, geometry exhibits the highest correlation with overall creativity, whereas texture shows the weakest alignment. This observation that structural or shape-based factors play a larger role in overall creativity perception aligns with findings in human visual perception, where coarse, global structural information is processed before fine-grained details~\cite{navon1977forest,marr2010vision}.
These results suggest that emphasizing geometric creativity may be particularly effective for achieving highly creative outcomes.

\section{Applications}

\subsection{Creativity Assessment}
\noindent\textbf{Representative Diffusion Models.}
As a reward model, CREward can provide type-wise creativity scores, which can be used to compare between generative models in terms of creative generation capability. 
We collect 100 object-agnostic creative prompts from LLMs and generate 10 images from five generative models used in Section \ref{sec:methodology}. 
We then compute and compare creativity scores from CREward in Figure~\ref{fig:model_comparison}.
Our results show that Hunyuan-DiT attains the highest creativity scores across all types. Conversely, Flux-schnell yields the lowest scores, consistent with the behavior of distillation-based models that often sacrifice diversity to improve fidelity.

\noindent\textbf{Creative Generation Methodologies.}
CREward can also serve as a benchmark for comparing and analyzing creative generation methodologies. For instance, Figure~\ref{fig:cre_method} illustrates the distribution of creativity scores on two object-oriented creative generation methods, ConceptLab~\cite{richardson2024conceptlab} and C3~\cite{han2025enhancing} compared with its baseline models.

As shown in Figure~\ref{fig:cre_method}, ConceptLab achieves the highest creativity in material and texture types. C3, on the other hand, demonstrates comparable or slightly enhanced geometry creativity for certain models, while consistently yielding substantial improvements over its baseline models across all creativity types. 
This way, CREward provides a structured and interpretable basis for consistent comparison across creative generation approaches, allowing their improvements relative to baseline models to be more clearly evaluated.

\subsection{Creative Sample Acquisition}
\noindent\textbf{Creative Sample Filtering.}
CREward can also be utilized to identify the most creative samples through large-scale assessment. We first generate 100 creative prompts from LLMs without specifying creativity types, and then produce 50 samples per prompt using various generative models. For each creativity type, we use the maximum sample score per prompt to identify the most and least creative prompts, as shown in Figure~\ref{fig:cre_filtering}. The Top 5 and Bottom 5 samples are selected from the 100 prompts for each type. As shown, samples with the highest scores clearly exhibit the corresponding type of creativity, whereas those with the lowest scores appear more ordinary and less distinctive.

\begin{figure*}[t!]
    \centering
    \includegraphics[width=\linewidth]{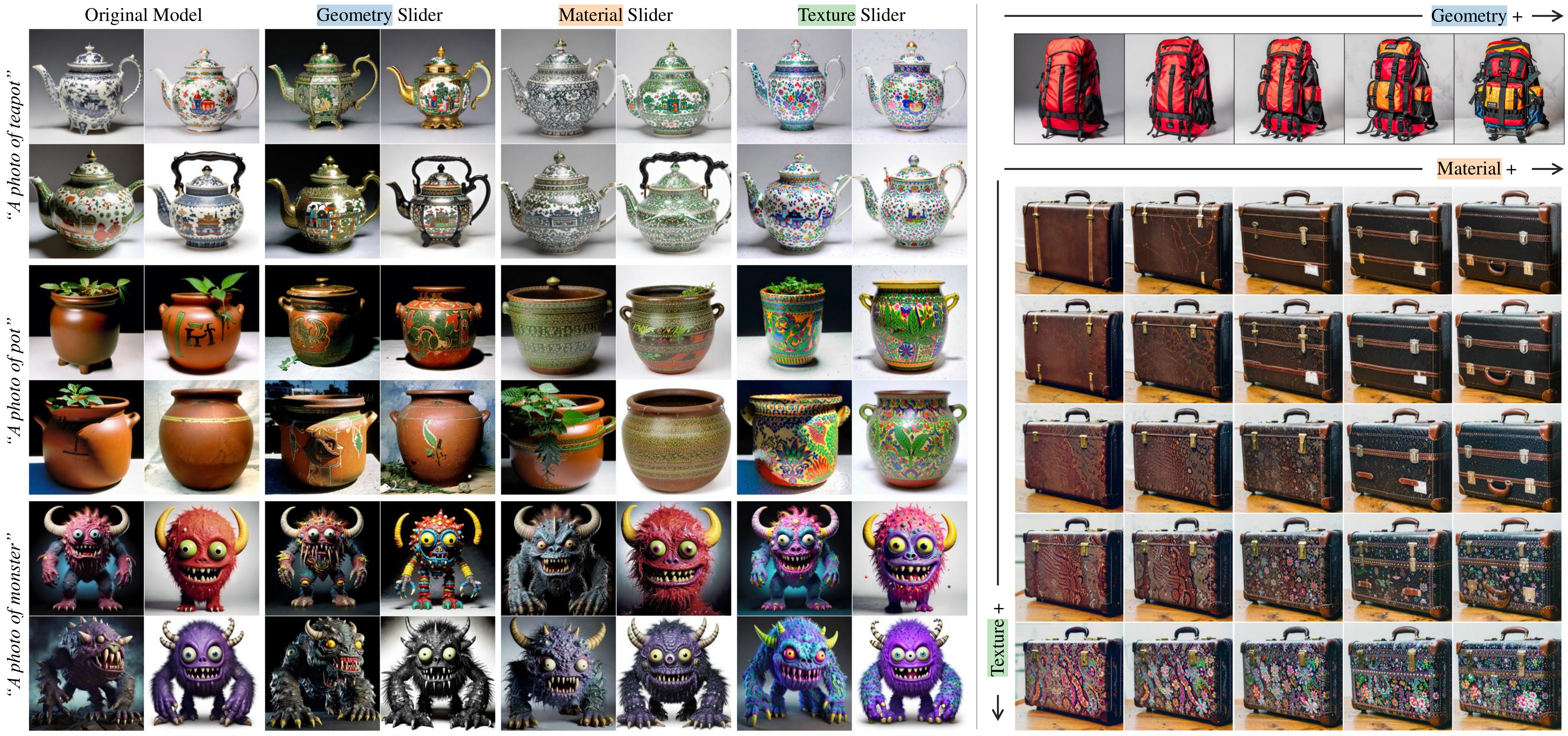}
    \caption{Type-specific sliders enhance the corresponding creative properties. (Left) Examples of guided generations for each type with LoRA weight $w=0.5$. (Right) Slider strength controls the extent of the effect, and multiple sliders can be mixed.}
    \label{fig:lora}
\end{figure*}

\noindent\textbf{Design Inspiration.}
Building on CREward’s ability to filter highly creative samples, the framework can naturally extend to human–AI co-creativity. By generating and evaluating a large pool of samples, we can efficiently surface the most creative examples and present them to product designers to support ideation and inspiration.

\begin{figure}
    \centering
    \includegraphics[width=1\linewidth]{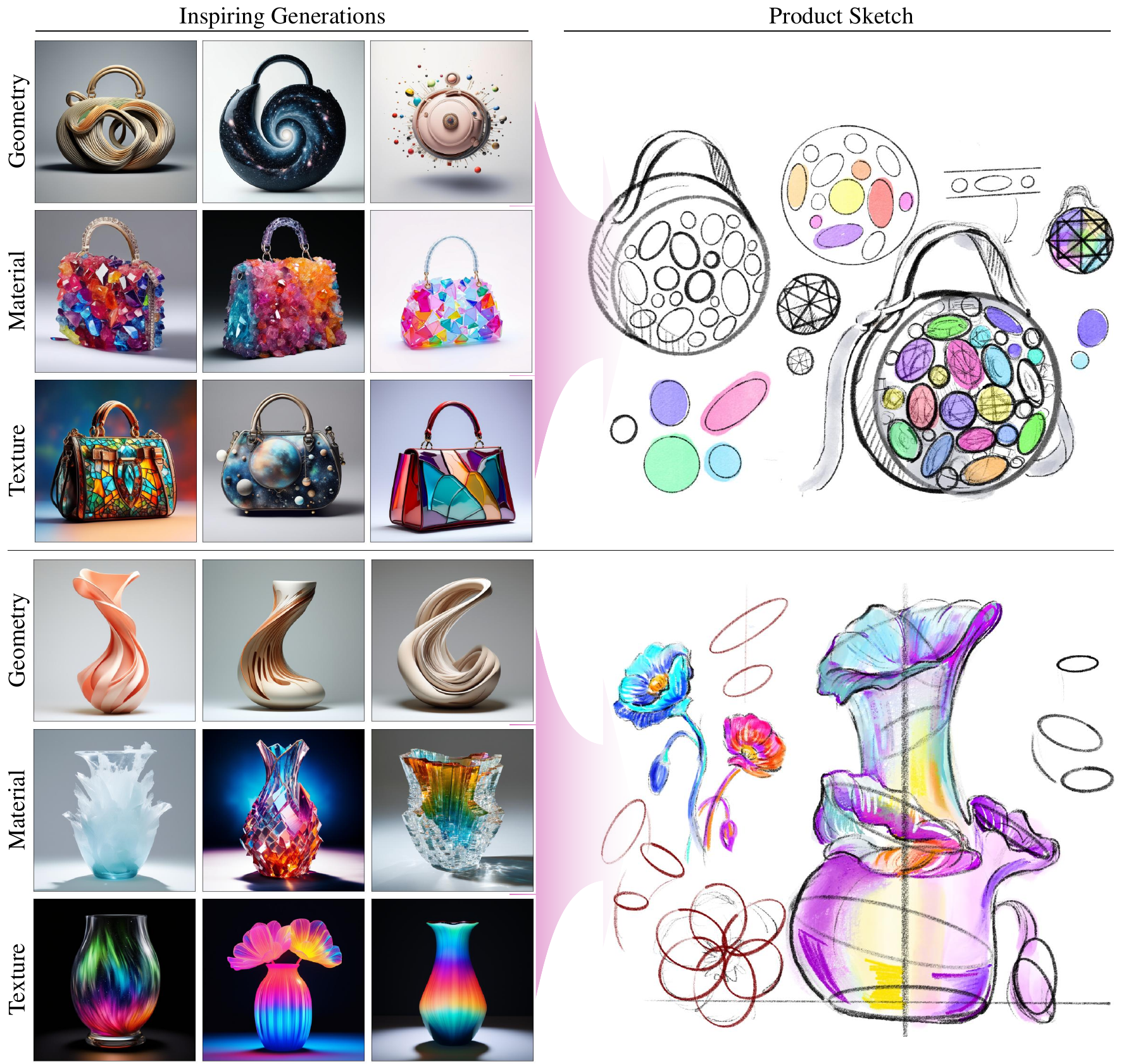}
    \caption{Human sketches and inspiring samples selected from the CREward inspiration set.}
    \label{fig:sketch}
\end{figure}

The examples in Figure~\ref{fig:sketch} illustrate such a scenario. From a collection of 1,500 images, we select the Top 30 images (Top 2\%) for each creativity type and provide them to an expert who has worked as a professional industrial designer. Figure~\ref{fig:sketch} shows the expert’s handbag and vase designs, which were inspired by specific AI-generated references from our automatic selection of inspiring set.
This demonstrates how our method can accelerate human–AI co-creation by reducing the burden of searching for creative references and offering diverse, type-specific creative samples that enrich the designer’s ideation process.

\noindent\textbf{Guiding Creative Generations.} 
To make the scores actionable, we fine-tune a pretrained T2I diffusion model to steer generation toward desired creative properties. We adopt a one-step extrapolation scheme to alleviate credit-assignment issues in iterative denoising \cite{xu2023imagereward,gandikota2025sliderspace}, and update only LoRA parameters for efficiency. 
We synthesize training data by generating 20 images per prompt $p$ of the form \textit{``a photo of \{obj\}''} for five objects (chair, vase, handbag, car, bowl) with the original model.
For the synthesized pairs \((\mathbf{x}_t,p,t)\) at time $t$, we optimize a reward-increasing objective coupled with the standard diffusion (noise-prediction) loss. Let \(\epsilon\sim\mathcal{N}(0,I)\) be the ground-truth noise, \(\epsilon_\phi(\mathbf{x}_t,p,t)\) the predicted noise, and \(\hat{\mathbf{x}}_{0,t}\) the one-step estimate of the clean sample obtained via DDPM-style inversion. For type \(c\) with reward model \(f^{(c)}_\theta(\cdot)\), we use the following formulation:
\begin{align}
\mathcal{L}_{\text{cre}} = -\,f^{(c)}_\theta\!\big(\hat{\mathbf{x}}_{0,t}\big), \;\;
\mathcal{L}_{\text{pre}} = \big\|\epsilon_\phi(\mathbf{x}_t,p,t)-\epsilon\big\|_2^2,
\end{align}
and therefore the total objective is:
\begin{align}
\mathcal{L} = \mathcal{L}_{\text{cre}} + \lambda\,\mathcal{L}_{\text{pre}}.
\end{align}
Additional training details are provided in Appendix~\ref{sec:sup_lora_setting}.
Finally, training yields type-specific CREward-LoRA \emph{sliders} (i.e., geometry, material, texture, overall) that can be applied at inference to control the corresponding creative property.

\begin{figure}[t!]
    \centering
    \includegraphics[width=1\linewidth]{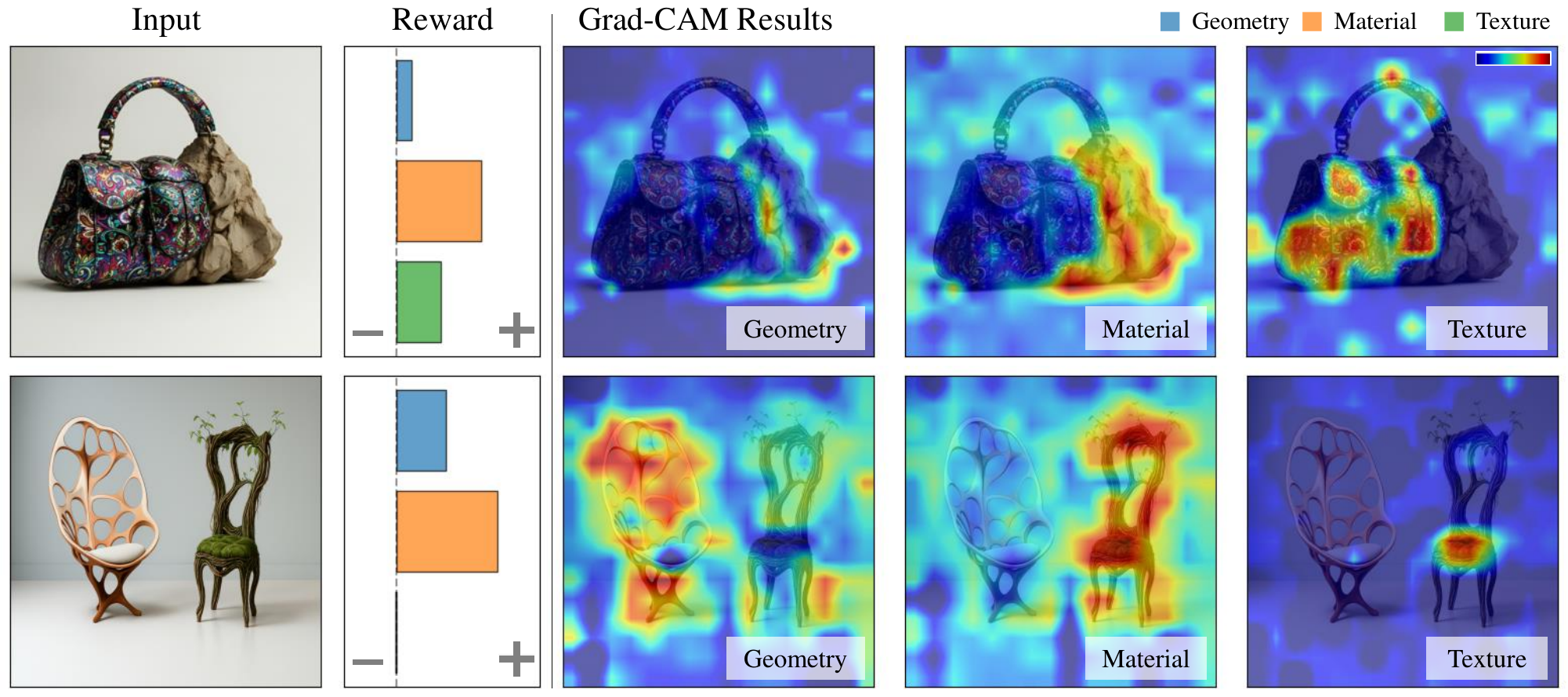}
    \caption{Grad-CAM results on creative images with combinations of various creativity types.}
    \label{fig:grad_cam_main}
\end{figure}
Figure~\ref{fig:lora} shows samples produced using the sliders for each creativity type. Attaching a slider selectively enhances the corresponding creative property even for target objects not seen during training; moreover, multiple sliders can be blended to induce several creativity types simultaneously. We provide additional experiments with LoRA sliders, including quantiative comparisons and various combinations, in Appendix~\ref{sec:sup_lora}. 

This approach enables diverse and controllable creative generation, providing targeted manipulation of creativity rather than relying on random or bias-prone outputs.

\subsection{Explainable Creativity}


Formulated as a differentiable function, CREward can be extended or combined with Explainable AI (XAI) methods to identify which parts of an object contribute to each type of creativity. For instance, we apply Grad-CAM \cite{selvaraju2017grad}, an input-attribution XAI technique, to visualize which pixels in the input image influence each type's CREward value.
Figure~\ref{fig:grad_cam_main} presents examples of Grad-CAM applied to CREward. Each input image exhibits different forms of creativity localized in different regions, and the Grad-CAM visualizations highlight the regions that correspond to each creativity type.
By integrating CREward with XAI methods, we open a research direction toward more systematic explanations and deeper understanding of creativity. See Appendix~\ref{sec:sup_gradcam} for more examples.

\section{Conclusions}
\label{sec:conclusion}
In this work, we introduce a framework for evaluating and guiding generative creativity by decomposing it into geometry, material, and texture.


We construct a human benchmark and analyze LVLM behavior, showing strong alignment with human creativity judgments, sometimes exceeding inter-human agreement. Building on this, we introduce \emph{CREward}, an open-weight reward model trained on LVLM-based annotations. CREward achieves high rank correlation with humans across all creativity types and outperforms other open models, demonstrating reliable creativity assessment without costly human labeling or proprietary LVLMs.

Our framework further supports practical applications, including benchmarking generative models, filtering highly creative samples, and enabling human–AI co-creation through diverse visual inspiration. With LoRA-based tuning, CREward scores become controllable signals for steering generation, while Grad-CAM offers interpretable insights into creativity-relevant regions.

However, our framework has limitations. Since the preference data assumes a minimum visual quality, CREward emphasizes novelty over value (Boden~\cite{boden1998creativity}), and may score semantically weak images highly (e.g., Figure~\ref{fig:cre_filtering}). When quality is uncertain, it should be paired with a value estimator (e.g., CLIP or BLIP-VQA) to balance novelty and value.

\begin{figure}[t!]
    \centering
    \includegraphics[width=1\linewidth]{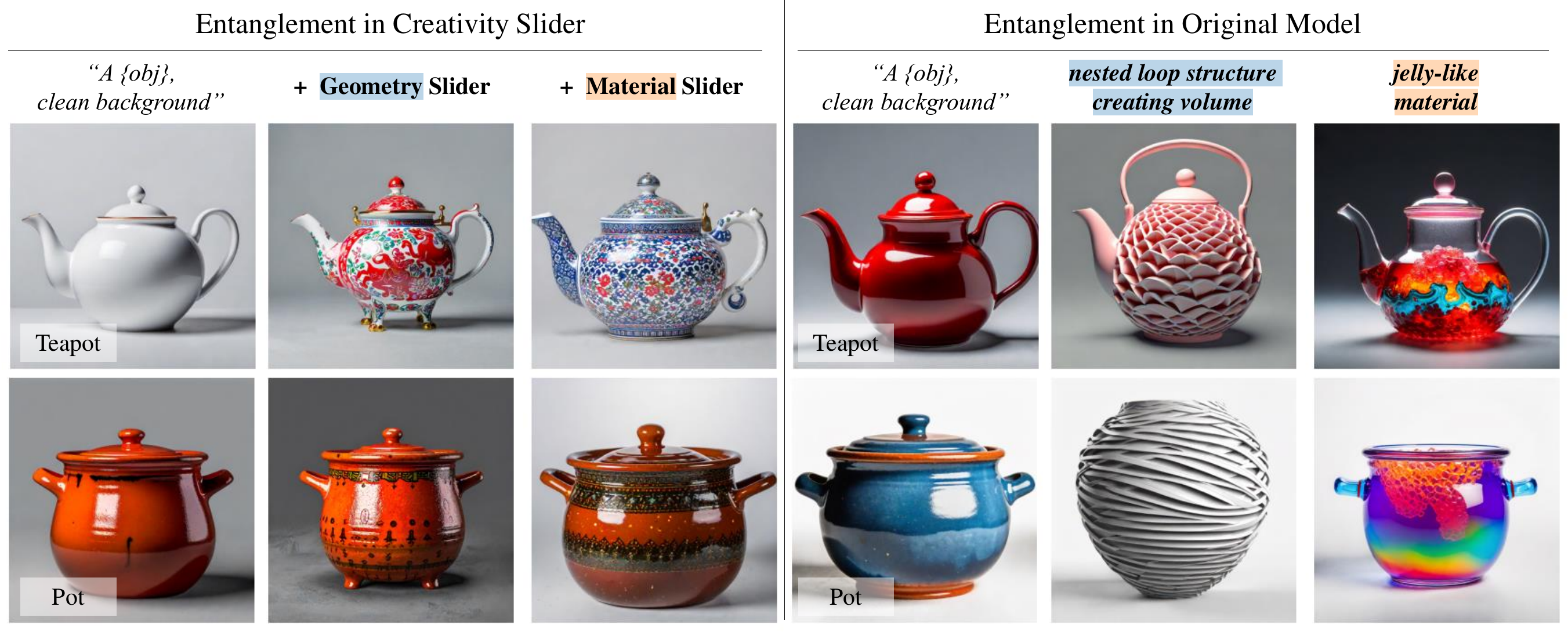}
    \caption{
            Illustration of entangled effects in creativity control using \textbf{(left)} LoRA sliders and \textbf{(right)} prompting. The columns on the right-hand side show that entanglement is inherent in the original model.. 
    }
    \label{fig:entangle}
\end{figure}


Another limitation is entanglement among creativity types: while LoRA sliders target specific types, cross-type effects are common (Figure~\ref{fig:entangle}), reflecting inherently entangled representations. Improving disentanglement for finer control is left for future work.

Further, while our approach generalizes to some unseen scenarios (see Appendix \ref{sec:sup_general}), extending it to multi-object settings and noisy environments remains future work.

Despite limitations, our work takes an initial step toward a measurable, interpretable framework for visual creativity by introducing a type-specific metric grounded in human perception. To our knowledge, this is the first to formalize creativity along geometry, material, and texture and provide a practical reward model. We hope this imperfect metric serves as a useful starting point for evaluating creative generation and inspiring future work.

\section*{Acknowledgments}
This work was supported by the Sogang University Research Grant of 2025 (202510003.01) and by gift funds from Adobe.

{
    \small
    \bibliographystyle{ieeenat_fullname}
    \bibliography{main}
}

\clearpage
\renewcommand{\thefigure}{\Alph{figure}}
\renewcommand{\thetable}{\Alph{table}}
\renewcommand{\thesection}{\Alph{section}}
\setcounter{page}{1}
\setcounter{figure}{0}
\setcounter{table}{0}
\setcounter{section}{0}
\onecolumn
\maketitlesupplementary

\section{Supplementary on CREBench}
\subsection{Selection of Types}
\label{sec:sup_types}
To identify the factors or elements that can enhance the creativity of an object, we first examine fundamental concepts from image formation. According to \citet{Hughes:2013:CGP}, for image formation, a geometric model is a model of something we plan to have appear in a picture, where it can be enhanced with various other attributes that describe the color or texture or reflectance of the materials involved in the model. Similarly, \citet{pharr2023physically} outlines 14 key base types in Physically Based Ray Tracer (pbrt), a widely used 3D rendering system. Among these, we identify Geometry (shape), Material, and Texture as types that physically influence an object’s creativity, whereas other factors, such as Camera or Light, may contribute to visual creativity without altering the object itself. 
Based on the literature, we select and specify types of creativity as in Figure~\ref{fig:types}.
\subsection{Creative Prompt Generation for LVLM Annotation Image Collection}
For CREBench, we generate prompts with an LLM, specifically \textit{ChatGPT-5}. 
For each creativity type (geometry, material, or texture), we collect 8 object-agnostic prompts and 12 object-specific prompts per object, respectively, allowing us to capture both global creativity and creativity conditioned on the properties of each object.

We first provide the LLM with the specifications of each creativity type and request 10 object-agnostic prompts per type, from which we manually select 8.
We then ask the LLM to generate 20 object-specific prompts per type, ensuring they do not overlap with the object-agnostic prompts, and manually select 12 from these.
The example prompts for object \textit{chair} are summarized in Table~\ref{tab:crebench_prompt}. Please note that the resulting prompts do not exhaust all possible forms of creative shape variation. Nonetheless, one can expand these prompts using the same recipe and produce more creative shapes for each category.
\vspace{1.4cm}

\begin{longtable}{c|c|cl}
\caption{CREBench prompts for each type for `chair'. We append \textit{``, clean background"} to all prompts.}
\label{tab:crebench_prompt}
\\
\toprule
 Type    & Obj- & No. & Prompt\\
\midrule

\multirow{20}{*}{Geometry} & \multirow{8}{*}{Agnostic} & 1 & a chair with a continuous flowing form made from a single curved surface \\
&&2 & a chair with asymmetrical legs and a twisted seat structure \\
&&3 & a chair with a spiral backbone-like frame supporting the seat \\
&&4 & a chair shaped from fluid ribbons forming seat and backrest in one motion \\
&&5 & a chair inspired by natural branching forms, like intertwined roots or coral \\
&&6 & a chair with an off-balance but visually stable structure \\
&&7 & a chair where seat and legs merge into one sculptural loop \\
&&8 & a chair composed of thin curved panels intersecting at elegant angles \\
\cline{2-4}
 & \multirow{12}{*}{Specific} & 9 & a chair with a double-layered shell that visually twists around the user \\
&&10 & a chair shaped like a flowing Möbius strip, single-surface design \\
&&11 & a chair with biomorphic form inspired by bone structures \\
&&12 & a chair built from modular cubes that can rearrange into different shapes \\
&&13 & a chair with creative spiral shape \\
&&14 & a chair with creative asymmetrical shape \\
&&15 & a chair with creative geometric shape \\
&&16 & a chair with creative organic shape \\
&&17 & a chair with creative floating shape \\
&&18 & a chair with creative animal-inspired shape \\
&&19 & a chair with creative modular shape \\
&&20 & a chair with creative multi-limbed shape \\
\hline
\multirow{20}{*}{Material} & \multirow{8}{*}{Agnostic} & 21 & a chair with creative liquid metal material that appears to flow but solidifies in form \\
&&22 & a chair with creative sand-textured material fused with transparent resin \\
&&23 & a chair with creative ceramic shard mosaic material \\
&&24 & a chair with creative iridescent shell-like material inspired by nacre \\
&&25 & a chair with creative paper-thin wood veneer material shaped fluidly \\
&&26 & a chair with creative aerogel-like ultra-light translucent material \\
&&27 & a chair with creative porous volcanic rock material \\
&&28 & a chair with creative layered silicone material creating gradient opacity \\
\cline{2-4}
 & \multirow{12}{*}{Specific} & 29 & a chair with creative sandstone and glass hybrid material \\
&&30 & a chair with creative rubberized foam material sculpted into precise geometry \\
&&31 & a chair with creative living moss-covered surface material \\
&&32 & a chair with creative mirror-polished ceramic composite material \\
&&33 & a chair with creative jelly-like material \\
&&34 & a chair with creative origami paper material \\
&&35 & a chair with creative molten glass material \\
&&36 & a chair with creative coral reef material \\
&&37 & a chair with creative knitted wool material \\
&&38 & a chair with creative translucent crystal material \\
&&39 & a chair with creative woven metal wire material \\
&&40 & a chair with creative recycled circuit board material \\
\hline
\multirow{20}{*}{Texture} & \multirow{8}{*}{Agnostic} & 41 & a chair with creative marbled ink color flowing organically \\
&&42 & a chair with creative tie-dye color inspired by fabric dye diffusion \\
&&43 & a chair with creative metallic gradient color fading from copper to teal \\
&&44 & a chair with creative translucent layered color giving a glassy depth \\
&&45 & a chair with creative color pattern inspired by topographic contour lines \\
&&46 & a chair with creative gradient stripes wrapping around curves \\
&&47 & a chair with creative woven color pattern resembling interlaced threads \\
&&48 & a chair with creative watercolor splash pattern, expressive and dynamic \\
\cline{2-4}
 & \multirow{12}{*}{Specific} & 49 & a chair with creative smoke-like swirling color, ethereal and soft \\
&&50 & a chair with creative mosaic color made of irregular tinted fragments \\
&&51 & a chair with creative heatmap-inspired color distribution, visually striking \\
&&52 & a chair with creative organic vein-like color pattern inspired by minerals \\
&&53 & a chair with creative galaxy-inspired color \\
&&54 & a chair with creative candy-swirled color \\
&&55 & a chair with creative rusted metallic color \\
&&56 & a chair with creative stained glass color \\
&&57 & a chair with creative fractal pattern \\
&&58 & a chair with creative floral-mechanical fusion pattern \\
&&59 & a chair with creative maze-like geometric pattern \\
&&60 & a chair with creative abstract pixel pattern \\
\hline
\bottomrule
\end{longtable}

\clearpage
\newpage
\subsection{LVLM-driven Preference Annotation}
\label{sec:sup_lvlm_anno}
\paragraph{LVLM Annotator Selection.}
In Table~\ref{tab:gemma12}, we observe that Gemini-2.5 shows strong alignment with average human rankings, surpassing even inter-human correlation, indicating that while human rankings may vary due to the inherent subjectivity of creativity, LVLMs can approximate an aggregated or generalizable human perception of creativity. However, Gemini-2.5 is a closed-source model and therefore unsuitable for scalable creativity assessment, analysis, or guidance.
Gemma-3-12B model (requiring approximately 20 GB of GPU memory in 16-bit precision), although achieving the second-highest rank correlations for geometry and texture, shows significantly low correlation for material, resulting in high variance across types and making it unreliable as an annotator.
Gemma-3-27B, on the other hand, achieves average rank correlations comparable to or slightly higher than inter-human correlation, with substantially lower variance across creativity types.
This suggests that at least a model of Gemma-3-27B quality, requiring roughly 46.4 GB of GPU memory in 16-bit precision, is needed to provide stable and reliable annotations, which in turn underscores the necessity of lightweight creativity assessment models.

\begin{table}[H]
  \caption{Rank correlation with human ranks. Numbers in parentheses indicate standard deviations over the objects. Boldface indicates the highest score, and underline indicates the second highest score.
  The \actionchip{blue model} denotes the open-source LVLM and \thingchip{red model} denotes the closed source LVLM. 
  }
  \label{tab:gemma12}
  \centering
  \begin{tabular}{@{}c|c|c|c|c@{}}
    \toprule
    Type & Inter-Human & \thingchip{Gemini-2.5 Flash} & \actionchip{Gemma-3-27b-it} & \actionchip{Gemma-3-12b-it}\\
    \midrule
    Geometry &  0.71 (0.07) & \textbf{0.80 (0.02)} & 0.56 (0.11) & \underline{0.57	(0.10)} \\
    \midrule
    
    Material & 0.63 (0.10) &  \textbf{0.75 (0.09)} & \underline{0.66 (0.10)} &  0.22	(0.30)\\
    \midrule
    
    Texture & 0.46 (0.10) & \textbf{0.74 (0.06)} & 0.60 (0.05) & \underline{0.73	(0.06)} \\
    \midrule
    
    Overall & 0.65 (0.10) &  \textbf{0.68 (0.04)} & 0.57 (0.08) & \underline{0.61	(0.08)} \\
    \midrule
    \midrule
    Average (over types) & 0.60 (0.13) & \textbf{0.76	(0.03)} & \underline{0.61	(0.05)} & 0.50	(0.26)\\	
    \bottomrule
  \end{tabular}
\end{table}

\begin{figure}[h!]
    \centering
    \includegraphics[width=0.8\linewidth]{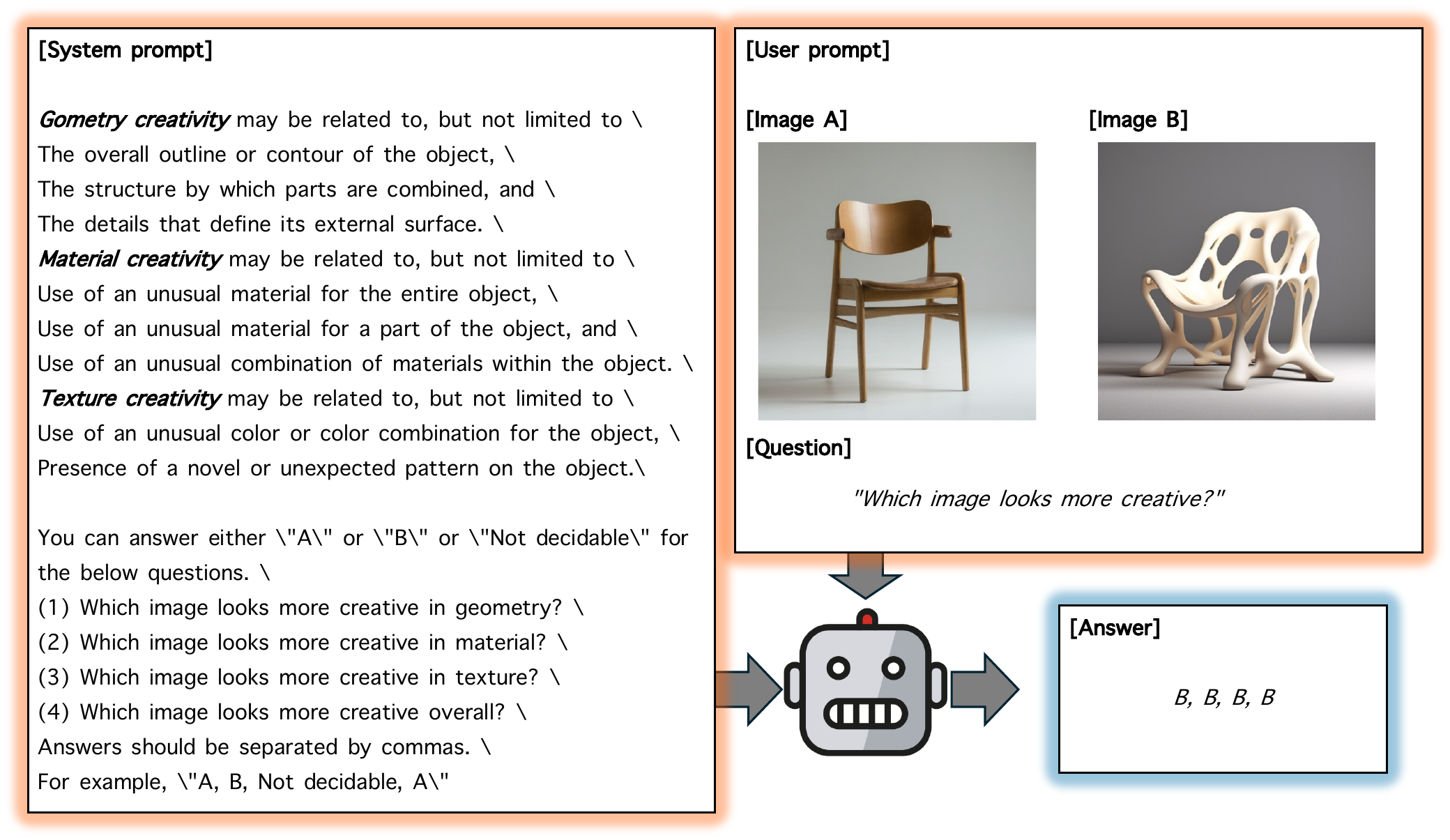}
    \vspace{-0.2cm}
    \caption{System prompt and user prompt provided to LVLM for preference annotation. }
    \label{fig:annotation}
\end{figure}

\paragraph{Query to LVLM for Preference Annotation.}
To annotate type-specific creativity preferences between two images, we first provide the LVLM with specifications for each creativity type, along with instructions regarding the questions and expected answer format. We then supply a pair of images generated from the prompts in Table~\ref{tab:crebench_prompt}, accompanied by a query asking which image is more creative for each type. The LVLM outputs its preference for each category. The overall annotation procedure is illustrated in Figure~\ref{fig:annotation}. 


\section{Supplementary on Creativity Assessment}
\subsection{LLM Generated 100 Creative Prompts}
\vspace{-0.5cm}
\begin{table}[b!]
\small
\centering
\caption{LLM generated creative prompts for creativity assessment and filtering.}
\vspace{-0.2cm}
\begin{tabularx}{\textwidth}{c|C|c|C}

\toprule
\textbf{No.}     & \Centering \textbf{Prompt} & \textbf{No.} & \Centering \textbf{Prompt}\\
\midrule
P1	&a {obj} with creative stained glass color	&P51	&a {obj} melting into the ground\\
P2	&a {obj} painted in shimmering holographic tones	&P52	&a {obj} split into mirrored halves\\
P3	&a {obj} with glowing neon gradient colors	&P53	&a {obj} morphing into another object\\
P4	&a {obj} designed in pastel watercolor blends	&P54	&a {obj} growing like a living plant\\
P5	&a {obj} with metallic rainbow surface	&P55	&a {obj} dissolving into particles\\
P6	&a {obj} covered in iridescent oil-slick color	&P56	&a {obj} with impossible infinite loops\\
P7	&a {obj} with glowing bioluminescent color	&P57	&a {obj} wrapped in glowing wires\\
P8	&a {obj} made of vibrant crystal shards	&P58	&a {obj} orbiting around itself\\
P9	&a {obj} with ultraviolet-inspired hues	&P59	&a {obj} with surreal shadow play\\
P10	&a {obj} in abstract Fauvist color	&P60	&a {obj} covered in galaxy patterns\\
P11	&a {obj} with intricate fractal pattern	&P61	&a {obj} with cosmic nebula colors\\
P12	&a {obj} decorated with swirling paisley motifs	&P62	&a {obj} featuring starry constellation patterns\\
P13	&a {obj} featuring geometric tessellation pattern	&P63	&a {obj} decorated with planetary textures\\
P14	&a {obj} with maze-like labyrinth pattern	&P64	&a {obj} infused with aurora borealis colors\\
P15	&a {obj} with Escher-style interlocking shapes	&P65	&a {obj} painted with lunar crater patterns\\
P16	&a {obj} covered in creative graffiti patterns	&P66	&a {obj} glowing like a miniature sun\\
P17	&a {obj} with mosaic tile surface	&P67	&a {obj} made of asteroid rock surface\\
P18	&a {obj} designed with mandala-inspired pattern	&P68	&a {obj} shaped like a spiral galaxy\\
P19	&a {obj} with origami-fold pattern	&P69	&a {obj} dissolving into stardust\\
P20	&a {obj} covered in tribal textile designs	&P70	&a {obj} in ancient Egyptian hieroglyph style\\
P21	&a {obj} with surreal liquid metal surface	&P71	&a {obj} with Mayan calendar pattern\\
P22	&a {obj} made of soft clouds	&P72	&a {obj} inspired by Aboriginal dot art\\
P23	&a {obj} constructed from glowing lava	&P73	&a {obj} covered in Celtic knot patterns\\
P24	&a {obj} made of translucent ice	&P74	&a {obj} decorated with Japanese ukiyo-e style\\
P25	&a {obj} carved from polished wood with intricate veins	&P75	&a {obj} featuring Islamic geometric patterns\\
P26	&a {obj} designed from woven bamboo	&P76	&a {obj} designed in medieval stained glass style\\
P27	&a {obj} covered in dripping paint texture	&P77	&a {obj} carved with runic symbols\\
P28	&a {obj} with cracked ceramic glaze	&P78	&a {obj} with Art Nouveau swirling lines\\
P29	&a {obj} crafted from woven threads	&P79	&a {obj} decorated in Bauhaus style blocks\\
P30	&a {obj} made of recycled paper pulp	&P80	&a {obj} glowing with bioluminescent algae patterns\\
P31	&a {obj} with asymmetrical surreal shape	&P81	&a {obj} with butterfly wing textures\\
P32	&a {obj} stretched and twisted into surreal curves	&P82	&a {obj} covered in peacock feather patterns\\
P33	&a {obj} folded into impossible geometry	&P83	&a {obj} with tiger stripe patterns\\
P34	&a {obj} with exaggerated oversized proportions	&P84	&a {obj} decorated in koi fish scales\\
P35	&a {obj} designed as a minimalistic flat silhouette	&P85	&a {obj} patterned like honeycomb\\
P36	&a {obj} with fractal-inspired branching shape	&P86	&a {obj} made of seashell spirals\\
P37	&a {obj} shaped like melting wax	&P87	&a {obj} covered in beetle iridescence\\
P38	&a {obj} with kaleidoscopic symmetry	&P88	&a {obj} decorated with coral reef forms\\
P39	&a {obj} built from modular cubes	&P89	&a {obj} with dragonfly wing transparency\\
P40	&a {obj} shaped like flowing ribbons	&P90	&a {obj} with rainbow stained metal effect\\
P41	&a {obj} with velvet soft texture	&P91	&a {obj} dripping with golden liquid\\
P42	&a {obj} covered in shimmering sequins	&P92	&a {obj} glowing with plasma energy\\
P43	&a {obj} with sandpaper roughness	&P93	&a {obj} sculpted from smoke\\
P44	&a {obj} covered in fuzzy moss	&P94	&a {obj} carved from amethyst crystal\\
P45	&a {obj} with glitter dust surface	&P95	&a {obj} constructed from shifting sand\\
P46	&a {obj} wrapped in silk cloth texture	&P96	&a {obj} emerging from shattered glass\\
P47	&a {obj} made of translucent jelly	&P97	&a {obj} sculpted from living vines\\
P48	&a {obj} covered in crystalline frost	&P98	&a {obj} wrapped in neon outlines\\
P49	&a {obj} with bumpy coral-like texture	&P99	&a {obj} disassembled into floating fragments\\
P50	&a {obj} floating in zero gravity	&P100	&a {obj} woven from glowing threads of light\\
\bottomrule
\end{tabularx}\label{tab:assessment_prompts}
\vspace{-0.5cm}
\end{table}

\clearpage
\newpage
\subsection{Creative Samples used for Creativity Assessment}
\begin{figure}[h!]
    \centering
    \includegraphics[width=1\linewidth]{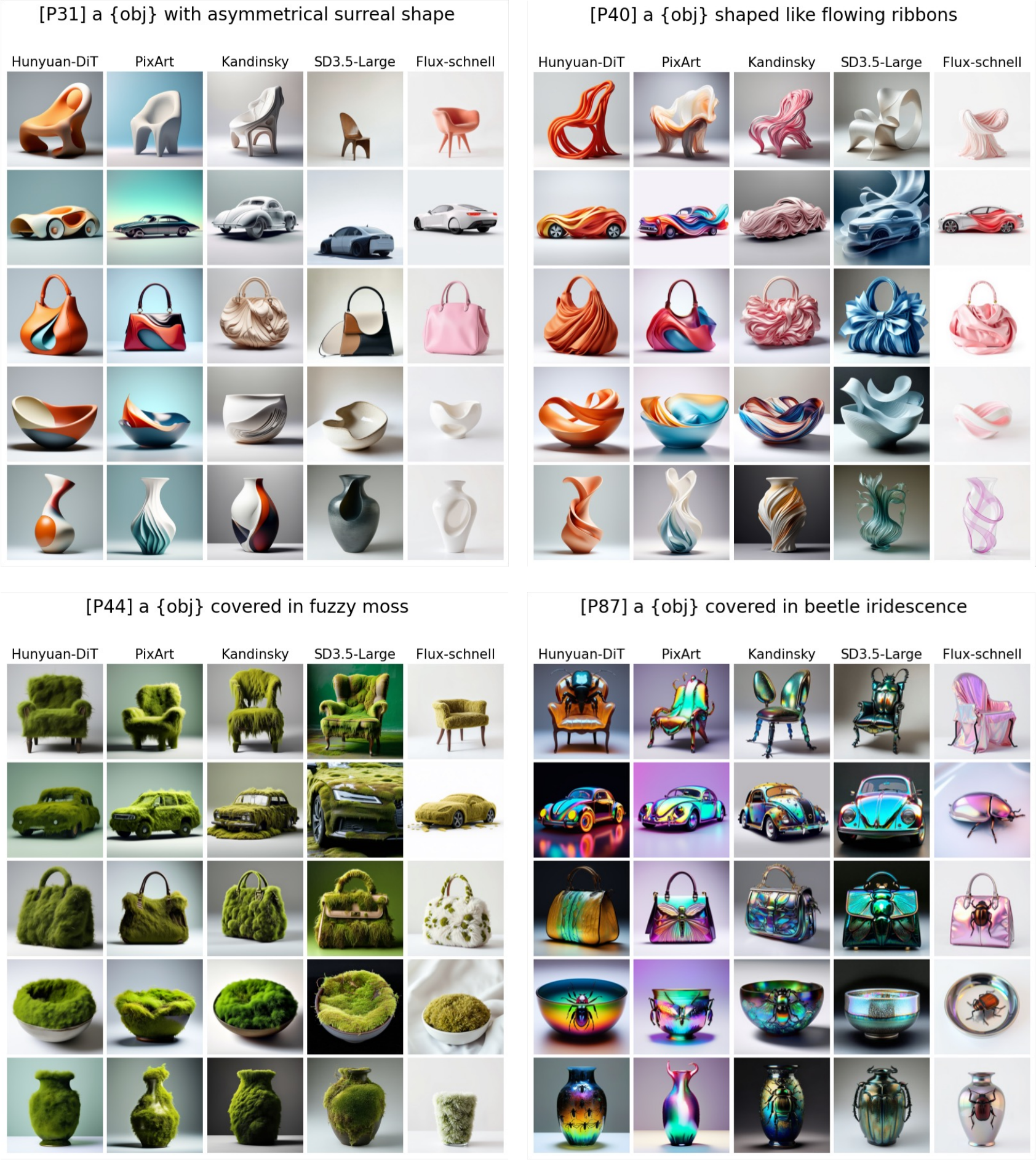}
    \caption{Examples of creative samples generated from the representative T2I generative models with the given prompt. The prompts are generated by ChatGPT-5. Full list of prompts can be found in Table~\ref{tab:assessment_prompts}.}
    \label{fig:grad_cam}
\end{figure}

\newpage
\section{Supplementary on CREward-LoRA}
\subsection{Creativity Slider}
\label{sec:sup_lora_setting}

After training CREward, we can assess creativity along the types, \textit{geometry, material, texture} as well as \textit{overall}. However, efficient inference alone has limited practical impact compared with the original LVLM-based evaluation.
To make the scores actionable, we fine-tune a pretrained T2I diffusion model to steer generation toward the desired creative properties. We adopt a one-step extrapolation scheme (to avoid credit-assignment issues in iterative denoising \cite{xu2023imagereward,gandikota2025sliderspace}) and update only LoRA parameters for efficiency 
We generate 20 images per prompt $y$ (\textit{"a photo of \{obj\}"}) for five objects (chair, vase, handbag, car, bowl) as the training data for LoRA in the original model.
For the synthesized training data, the training objective is formulated to increase the reward score of target type with the standard diffusion (i.e., noise-prediction) loss.
We additionally stabilize training with the standard diffusion (noise-prediction) loss.
Let $\textbf{x}_t$ be the noisy latent at step $t$, $\epsilon\sim N(0,I)$ the ground-truth noise, $\epsilon_\phi(\textbf{x}_t,y,t)$ the models' predicted noise, $\bar{\alpha}_t$ the scheduler's cumulative product, and $D(\cdot)$ the decoder. We form a one-step estimate of the clean latent
\begin{align}
    \bar{\textbf{x}}_{0,t}\leftarrow \frac{\textbf{x}_t-\sqrt{1-\bar{\alpha}_t}\epsilon_\phi(\textbf{x}_t,y,t)}{\sqrt{\bar{\alpha_t}}}
\end{align}
For type $c$ with reward model $f_\theta^{(c)}(\cdot)$, the losses are
\begin{align}
    L_{cre}=-f^{(c)}_\theta(D(\bar{\textbf{x}}_{0,t})),\;\; L_{pre}=||\epsilon_\phi(\textbf{x}_t,y,t)-\epsilon||_2
\end{align}
and the total objective is
\begin{align}
    L=L_{cre}+\lambda L_{pre}.
\end{align}
where $t$ is timestep, $\textbf{x}_t$ is corrupted sample. 
We run experiments on the SDXL-DMD2~\cite{yin2024onestep,yin2024improved} with 4 denoising steps. We set $\lambda=0.1$ and LoRA rank/alpha $r{=}8, \alpha{=}8$, respectively. We train type-specific LoRA sliders on these generations and then apply the trained sliders to unseen objects. Each slider is trained for 35 epochs on an A6000 (50\,GB) GPU with a batch size of 1 with learning rate $lr=1e^{-4}$ and 10 gradient accumulations.
Finally, training yields type-specific LoRA sliders (e.g., geometry, material, texture, overall) that can be applied at inference to control the corresponding creative property.

\clearpage
\newpage
\subsection{Experimental Results on CREward-LoRA}
\label{sec:sup_lora}
\subsubsection{Settings}

We provide additional experiments of creativity-guided sampling using our sliders or using prompt variants. For visualization, we set slider strengths to \(w \in \{0.5, 0.7, 1.0\}\). For prompting, the \emph{prompt} reference is chosen at random from a pool of candidate prompts in Table~\ref{tab:5prompts}. 
Figure~\ref{fig:lora_qual_all} in Sec.~\ref{sec:sup_lora_qual} shows examples from our sliders and from prompting, for each creativity type. 

We also quantitatively evaluate our sliders and prompt variants under a creativity-focused protocol. For each creativity type, we adopt five creativity-guided prompts (Table~\ref{tab:5prompts}) as a supervised baseline, and prepare a set of 20 object categories (Table~\ref{tab:object_candidates}) and 20 random latent codes. We then generate images per condition, \emph{original} (no guidance), each \emph{slider}, and each \emph{prompt} baseline, using identical latent codes.
For every generated image, we report: (1) the CREward score (higher is better), and (2) the LVLM \emph{improvement ratio}, i.e., the fraction of pairwise comparisons whether the guided image is preferred over the original by the LVLM. We also assess \emph{consistency} with the original image by computing Euclidean and cosine distances in the SigLIP feature space (lower distance indicates higher consistency) and report in the following section. 

\begin{table}[h!]
\centering
\caption{Creativity-guided prompts for each type}
\begin{tabular}{lcl}
\toprule
 Type    &No. & Prompt Candidates\\
\midrule

Geometry & \begin{tabular}[c]{@{}c@{}}
C1 \\ C2 \\ C3 \\ C4 \\ C5
\end{tabular}& \begin{tabular}[c]{@{}l@{}}
``a \{obj\} with creative spiral shape, clean background''\\
``a \{obj\} with creative asymmetrical shape, clean background''\\
``a \{obj\} with creative geometric shape, clean background''\\
``a \{obj\} with creative organic shape, clean background''\\
``a \{obj\} with creative animal-inspired shape, clean background''
\end{tabular} \\
\midrule
Material & \begin{tabular}[c]{@{}c@{}}
C6 \\ C7 \\ C8 \\ C9 \\ C10
\end{tabular}& \begin{tabular}[c]{@{}l@{}}
``a \{obj\} with creative jelly-like material, clean background''\\
``a \{obj\} with creative molten glass material, clean background''\\
``a \{obj\} with creative coral reef material, clean background''\\
``a \{obj\} with creative translucent crystal material, clean background''\\
``a \{obj\} with creative woven metal wire material, clean background''
\end{tabular} \\
\midrule
Texture  & \begin{tabular}[c]{@{}c@{}}
C11 \\ C12 \\ C13 \\ C14 \\ C15
\end{tabular}& \begin{tabular}[c]{@{}l@{}}
``a \{obj\} with creative galaxy-inspired color, clean background''\\
``a \{obj\} with creative candy-swirled color, clean background''\\
``a \{obj\} with creative stained glass color, clean background''\\
``a \{obj\} with creative maze-like geometric pattern, clean background''\\
``a \{obj\} with creative abstract pixel pattern, clean background''
\end{tabular} \\
\bottomrule
\end{tabular}\label{tab:5prompts}
\end{table}

\begin{table}[h!]
\centering
\caption{Object candidates for quantitative comparison}
\begin{tabular}{c}
\toprule
Object Candidates\\
\midrule
`airplane', `backpack', `bed', `bench', `bicycle', `boat', `bookshelf',\\
`bus', `cup', `kite', `motorcycle', `scissors', `spaceship', `table', \\
 `teapot', `teddy bear', `toothbrush', `train', `truck', `umbrella'\\
\bottomrule
\end{tabular}\label{tab:object_candidates}
\end{table}


\clearpage
\newpage
\subsubsection{Quantitative Results}
\label{sec:sup_lora_quant}
\begin{figure}[h!]
    \centering
    \includegraphics[width=0.8\linewidth]{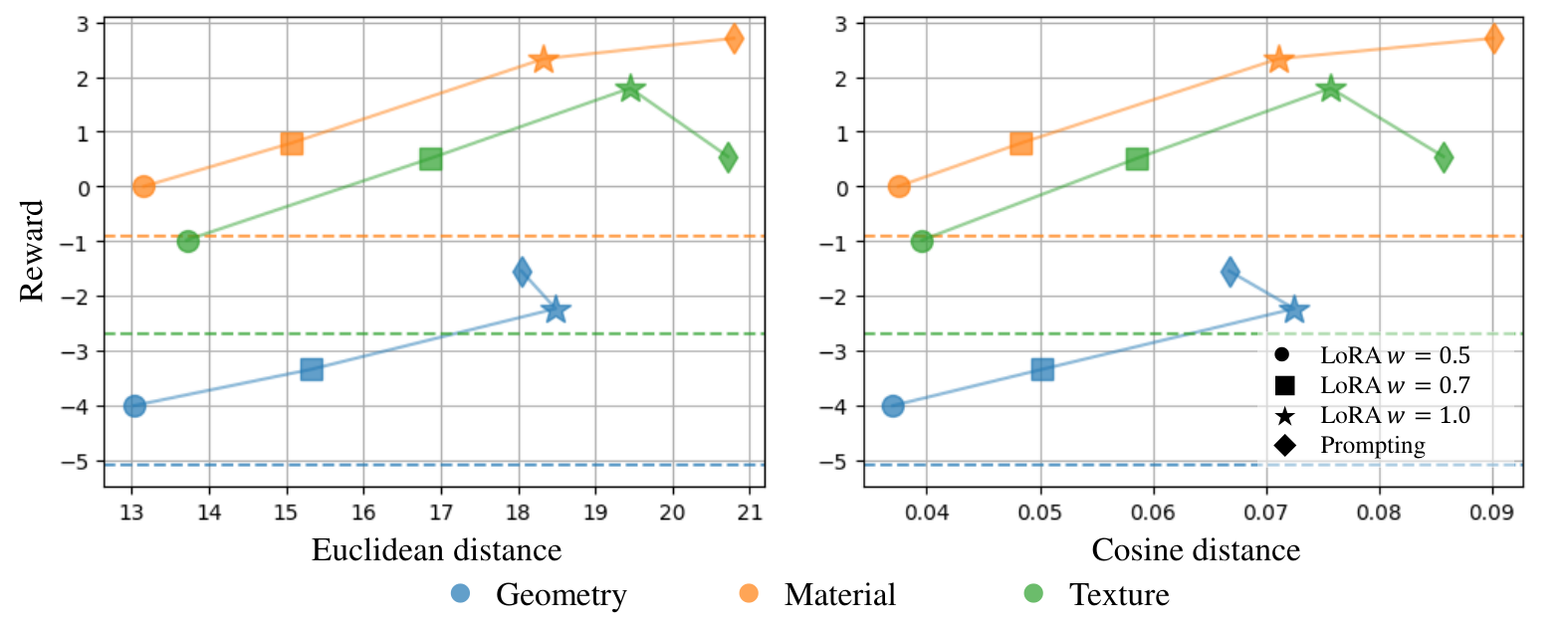}
    \caption{Average CREward score across object categories for each creativity type. Markers indicate creativity-guided methods; colors denote creativity types. The dashed line shows the average score of the original (unguided) generations.}
    \label{fig:lora_eval}
\end{figure}

We conduct quantitative comparisons based on generations produced without CREward-LoRA, with CREward-LoRA, and via direct prompting, all under the same random latent code.
Figure~\ref{fig:lora_eval} shows the CREward values versus distances for each CREward-LoRA configuration with different strengths (0.5, 0.7, 1.0) and for prompting, averaged over the 20 objects listed in Table~\ref{tab:object_candidates}.
CREward provides a type-specific creativity signal, whereas Euclidean and Cosine distances in SigLIP feature space indicate the deviation from the original identity.
For geometry, prompting achieves high rewards with less alteration of the object’s identity.
For material, prompting yields slight improvements in reward but introduces substantial changes, resulting in weaker identity preservation.
For texture, CREward-LoRA achieves higher creativity scores with noticeably smaller feature shifts, demonstrating that it can enhance creativity while preserving the original identity more effectively.

\begin{figure}[h!]
    \centering
    \includegraphics[width=0.8\linewidth]{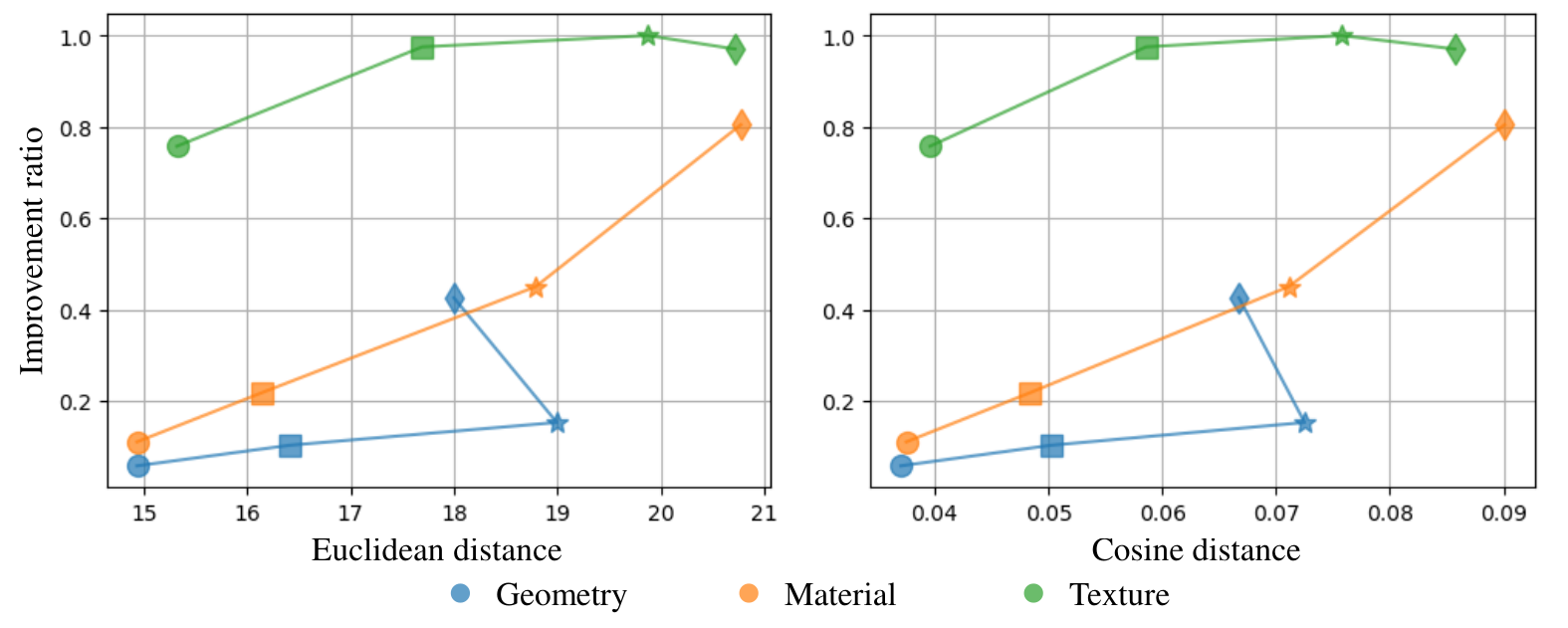}
    \caption{Average CREward score across object categories for each creativity type. Markers indicate creativity-guided methods; colors denote creativity types. The dashed line shows the average score of the original (unguided) generations.}
    \label{fig:lora_pref_eval}
\end{figure}

Figure~\ref{fig:lora_pref_eval} reports the improvement ratio for CREward-LoRA and prompting. To measure improvement, we ask the LVLM (Gemma-3) to choose the more creative image for each type between the original output and its creativity-enhanced counterparts (either CREward-LoRA or prompting), allowing for a \textit{tie} option. If the enhanced image is preferred, we treat it as an improvement. We then compute the improvement ratio as the proportion of improved samples out of 20 examples per object.

Consistent with Figure~\ref{fig:lora_eval}, prompting yields stronger improvements with less identity alteration for \textit{geometry}, and stronger improvements with larger identiy alteration for \textit{material}, whereas CREward-LoRA achieves close-to-perfect improvement ratio while requiring significantly less deviation from the original identity for \textit{texture}.


\clearpage
\newpage
\subsubsection{Qualitative Results}
\label{sec:sup_lora_qual}
\begin{figure}[h!]
    \centering
    \includegraphics[width=0.9\linewidth]{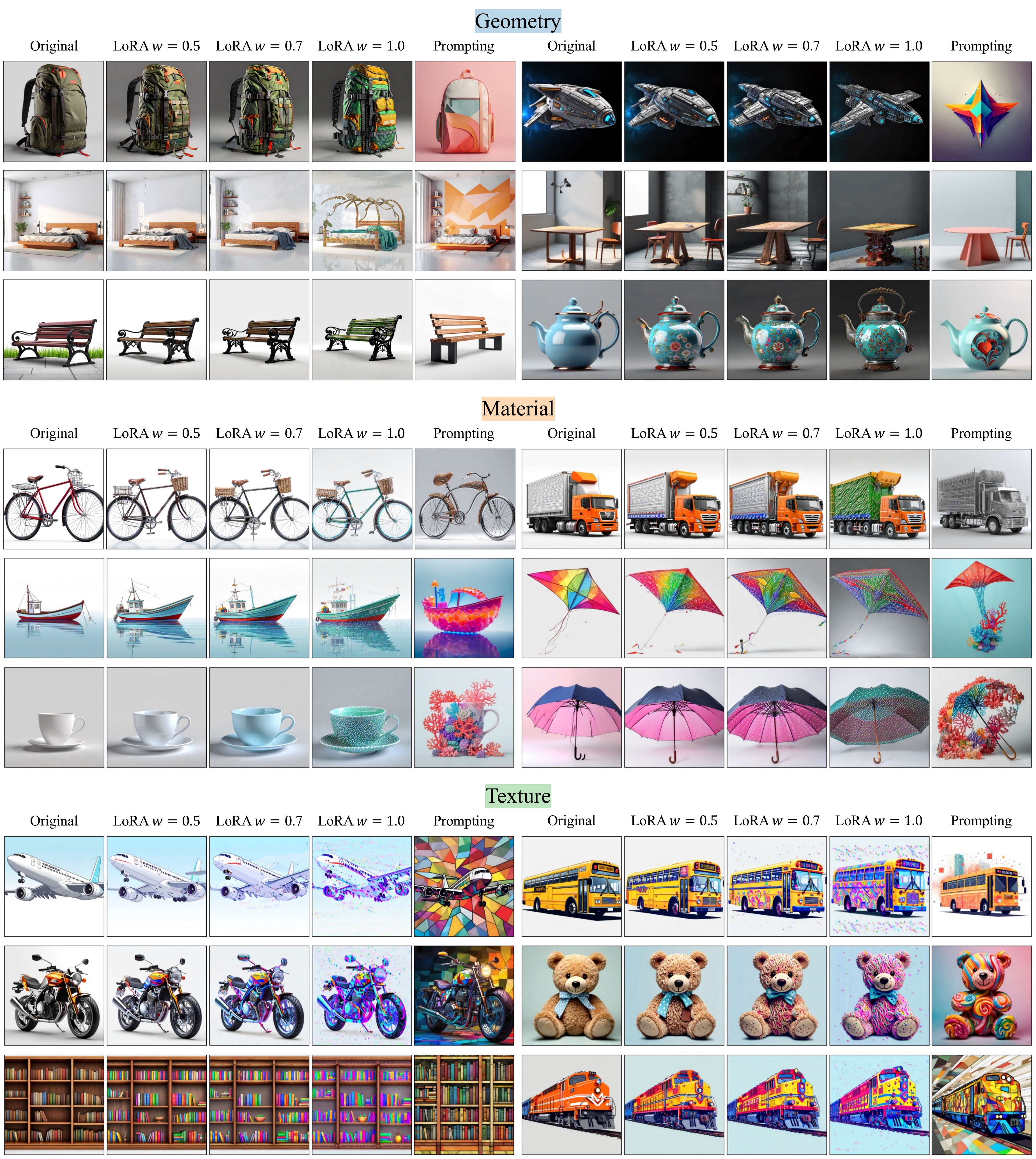}
    \caption{Randomly selected examples of creative generations for each type using our creativity slider and prompt control. For each case, the prompting image is randomly sampled from the candidate set.}
    \label{fig:lora_qual_all}
\end{figure}

\paragraph{Examples used in Quantitative Results.}
Figures~\ref{fig:lora_qual_all} show some samples generated from SDXL-DMD2 without CREward-LoRA (Original), with CREward-LoRA with various strengths ($w\in[0.5,0.7,1.0]$), and prompting. 
In \textit{Prompting} column, we randomly select among 5 prompts (listed in Table~\ref{tab:5prompts}) for each type.
Overall, while prompting can yield highly creative results, it typically introduces substantial changes to the original identity, even when using the same random latent code. In contrast, CREward-LoRA tries to improve the target creativity type while better preserving the original identity. Notably, CREward-LoRA also offers a controllable trade-off between the degree of alteration and the extent of creativity enhancement. 

\begin{figure}[h!]
    \centering
    \includegraphics[width=1\linewidth]{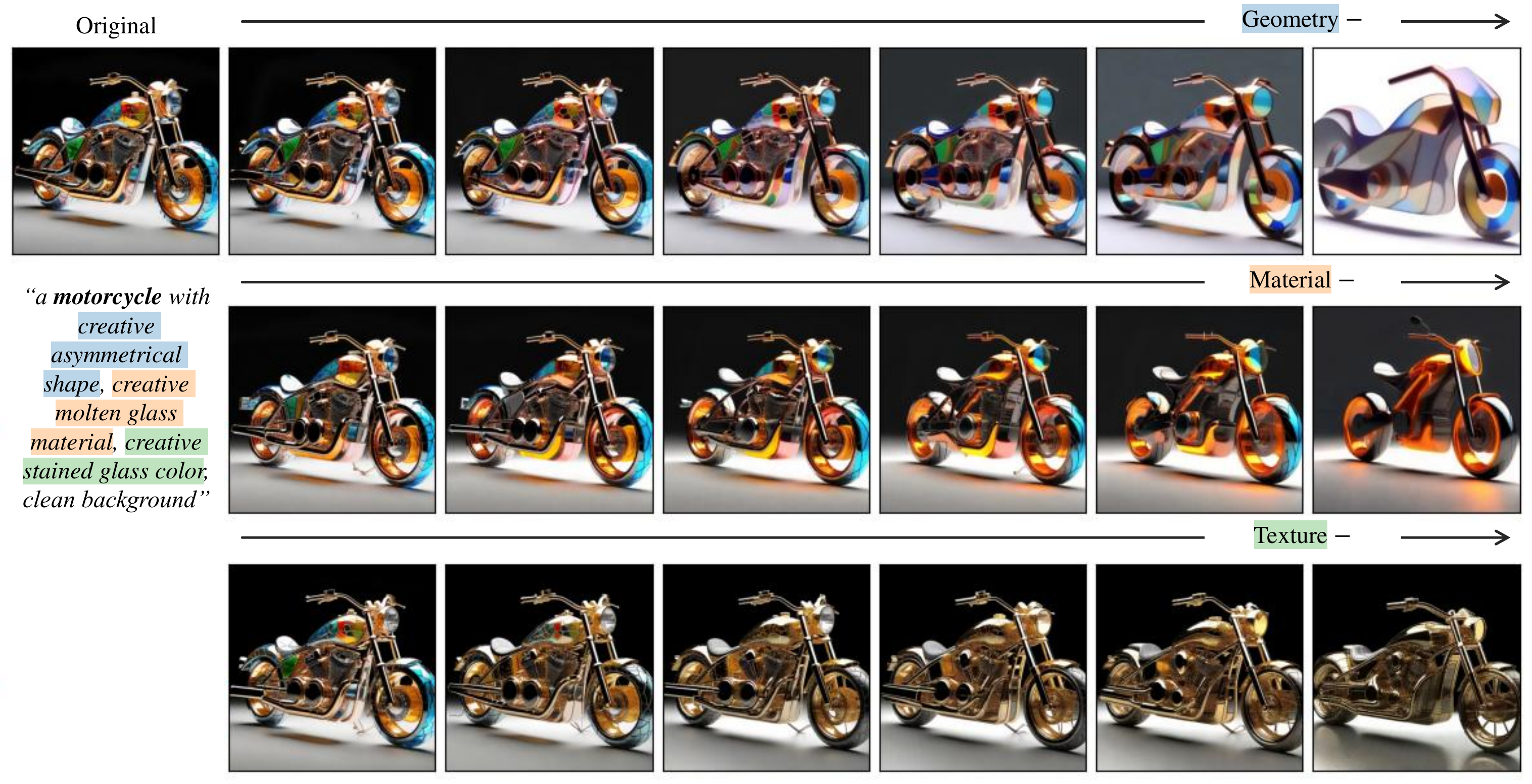}
    \caption{Applying CREward-LoRA in the negative direction to the same random latent code as the leftmost original output.}
    \label{fig:lora_minus}
\end{figure}

\paragraph{Adapting CREward-LoRA in Negative Direction.}
To demonstrate its effectiveness, we further evaluate whether the trained CREward-LoRA correctly captures each creativity type by applying it in the negative direction. Figure~\ref{fig:lora_minus} presents the results. Starting from the original output (top left), applying CREward-LoRA for the geometry type in the negative direction reduces joints and structural details while largely preserving colors and the steel-like material.
Applying the material slider in the negative direction preserves the overall shape but removes material richness, producing a flattened, textureless 3D appearance.
Finally, applying the texture slider negatively removes colorful patterns and yields a more monotone surface.

These results indicate that CREward-LoRA has indeed learned the underlying concept of each creativity type. This opens promising opportunity for future research on isolating and manipulating creativity types by steering CREward-LoRA through mixtures of positive and negative directions.

\begin{figure}[h!]
    \centering
    \includegraphics[width=1\linewidth]{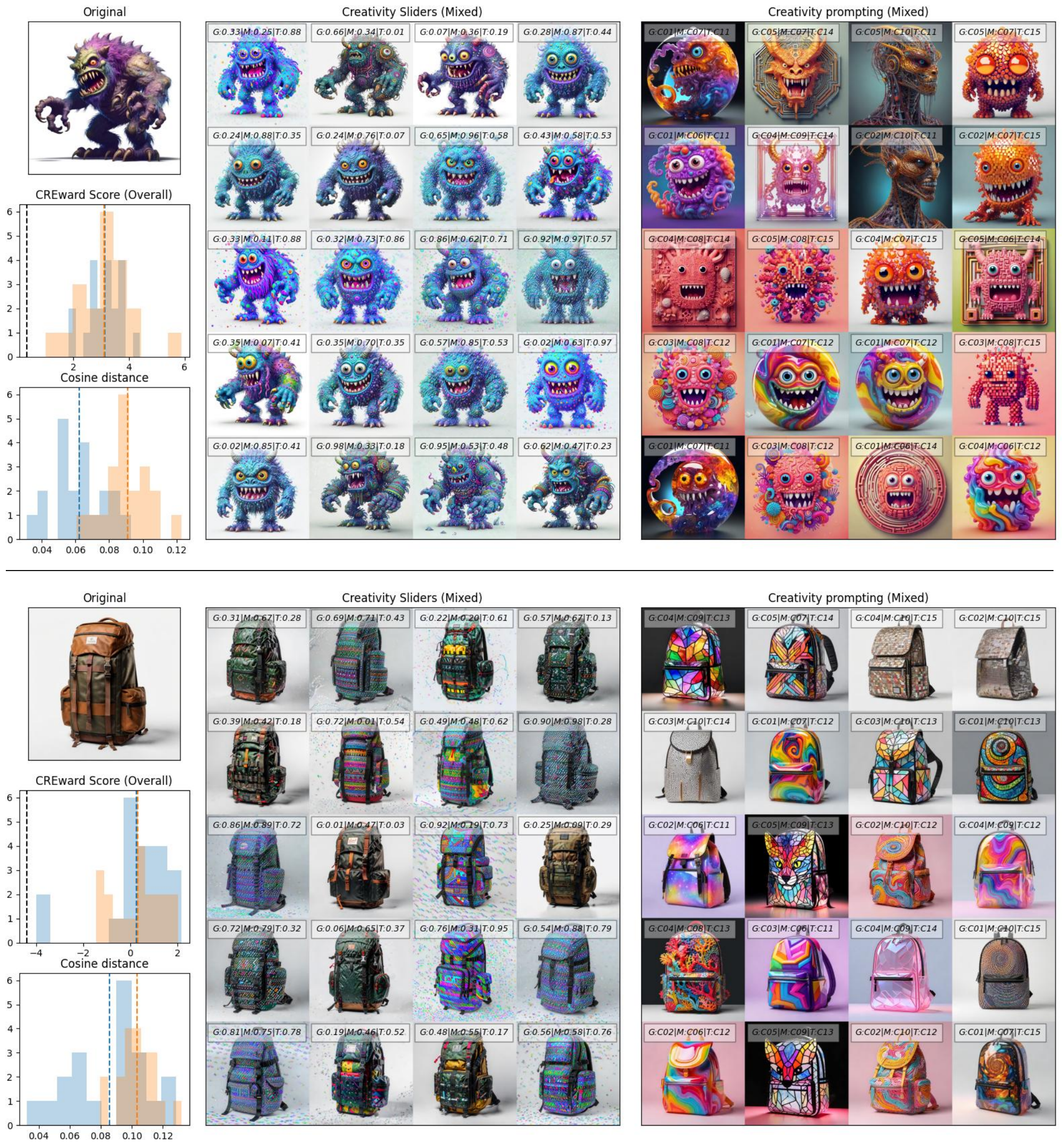}
    \caption{Results of random mixtures of creativity types for CREward-LoRA (second column) and prompting (third column). Each sample in these columns is generated using the same random latent code as the original output shown at the top left. The strength configuration for each type (for CREward-LoRA) or the prompts used for each type (for prompting) is indicated in the top-left text box of each sample.
In the leftmost column, the middle plot shows the overall CREward score distributions for CREward-LoRA (blue) and for prompting (orange). Colored dashed lines denote the respective means, and the black dashed line marks the CREward score of the original output. The bottom plot shows histograms of cosine distances between the original and enhanced outputs for both methods.}
    \label{fig:lora_cons}
\end{figure}

\paragraph{Random Cross-Type Combinations}
Figure~\ref{fig:lora_cons} illustrates CREward-LoRA’s ability to produce finely controlled mixtures of creativity types. Using the same random latent code as the original output (top left), we apply various strength configurations across types.
For prompting, since it cannot control the degree of each type, we randomly select one prompt per type from Table~\ref{tab:5prompts} and concatenate them to create mixed-type prompts, again using the same random latent code as the original output.

In the leftmost column, the middle figure shows the overall CREward distributions, blue for CREward-LoRA and orange for prompting. Colored dashed lines denote their respective means, and the black dashed line indicates the CREward score of the original output (top). The bottom figure in the leftmost column shows the histograms of cosine distances between the original and enhanced outputs for both methods.

These plots demonstrate that while both CREward-LoRA and prompting substantially increase creativity, CREward-LoRA achieves this with smaller feature-distance shifts, suggesting better preservation of the original identity. The bag examples in Figure~\ref{fig:lora_cons} illustrate this more clearly. While prompting often changes the subcategory of the original backpack, CREward-LoRA produces variations that maintain the original style while introducing controlled creative modifications.
Furthermore, the examples in the second column show that CREward-LoRA enables arbitrary mixing of creativity types with controllable strength, whereas prompting provides less control over the extent and interaction of mixed types.

\clearpage
\newpage
\section{Generalization on Unseen Data}
\label{sec:sup_general}
\begin{figure}[h]
    \centering
    \includegraphics[width=1\linewidth]{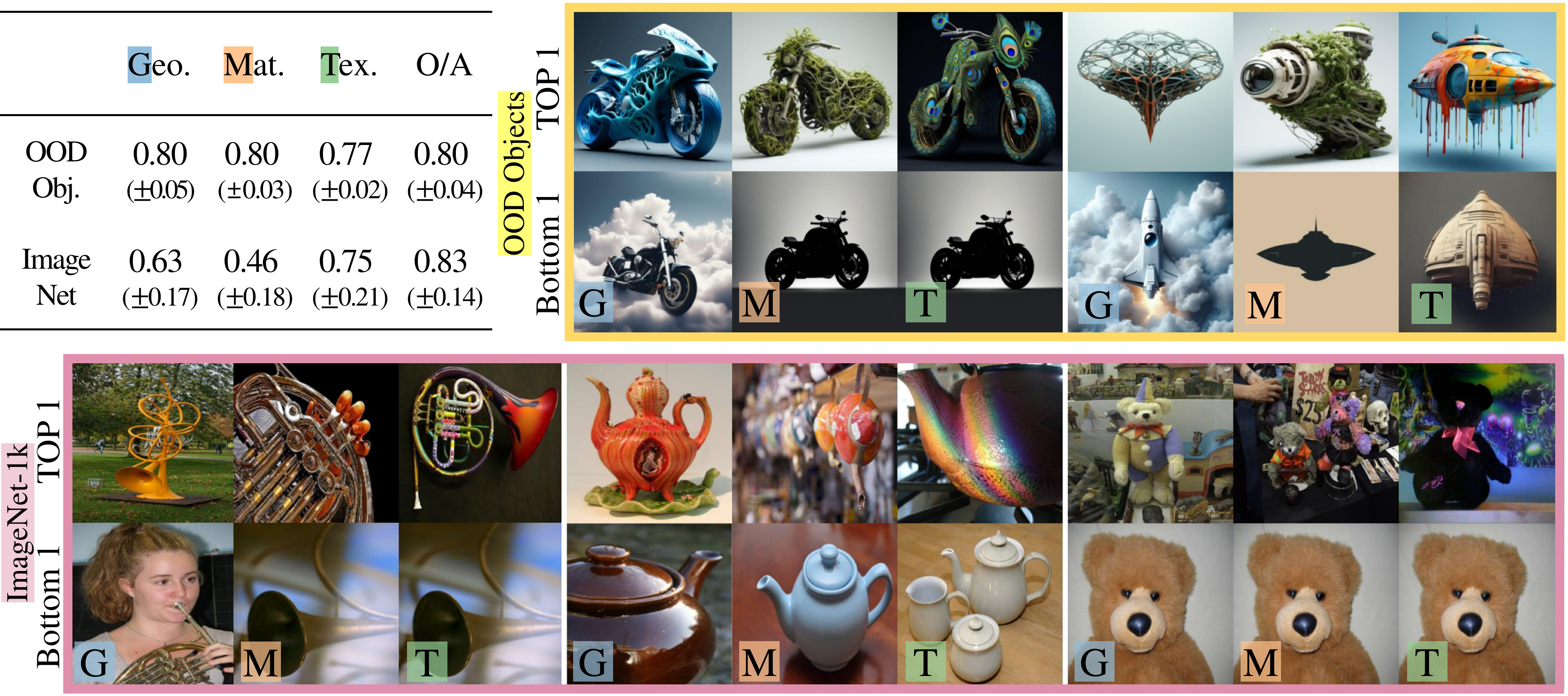}
    \vspace{-0.75cm}
    \caption{Experimental results on unseen scenarios.}
    \label{fig_rebuttal:imagenet_qual}
    \vspace{-0.67cm}
\end{figure}

We conduct additional experiments on (1) \colorbox{yellow!50}{\textbf{out-of-domain}} (OOD) synthesized images using five unseen objects, and (2) \colorbox{purple!20}{\textbf{real-world}} images from ImageNet-1k training set, focusing on 40 object-centric classes (e.g., horn). Figure~\ref{fig_rebuttal:imagenet_qual} reports preference alignments between CREward and Gemma. Alignments on OOD data are comparable to those reported in our paper. 
The ImageNet-1k training set contains approximately 1{,}300 images per class and is not curated for creative content, explaining lower alignment, but our method can help uncover relatively more creative samples within these classes, as the qualitative results suggest.

For (1), we generate 5k images using the same settings as in Sec.~3.1 and evaluate $n=1,000$ randomly sampled pairs, while for (2), we use $n = 15 \times 15$ exhaustive pairs between the top 15 and bottom 15 ranked images.

\clearpage
\newpage
\section{Additional Examples on GradCAM}
\label{sec:sup_gradcam}
\begin{figure}[h!]
    \centering
    \includegraphics[width=0.95\linewidth]{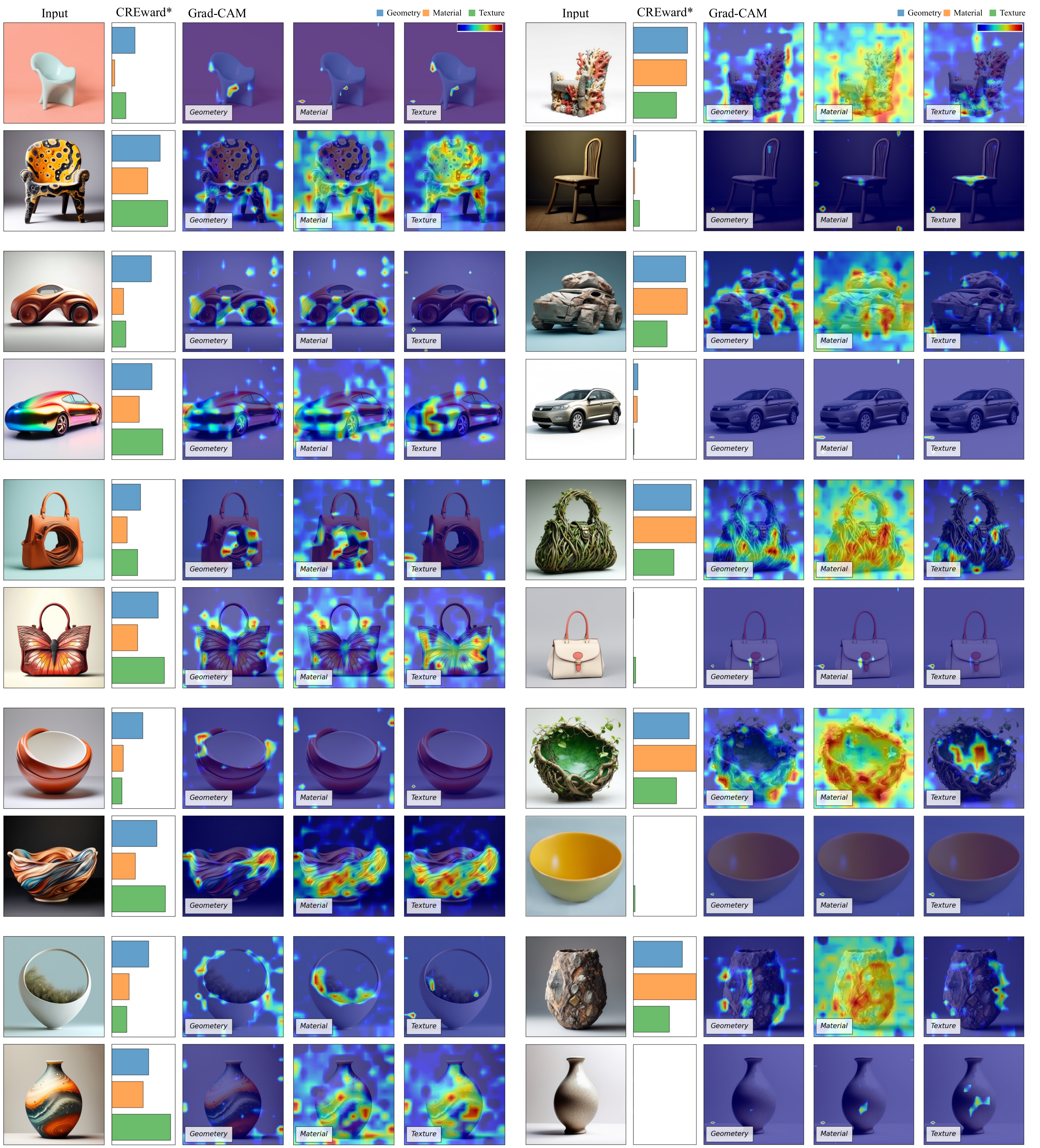}
    \caption{Additional Grad-CAM results on creative images with combinations of various creativity types. CREward* refers to min-max normalized CREward values on each type (blue: geometry, orange: material, green: texture), over 25 human evaluation benchmark for each object.}
    \label{fig:sup_grad_cam}
\end{figure}

\end{document}